\documentclass{article} 
\usepackage{paper_style,times}
\iclrfinalcopy

\usepackage{amsmath,amsfonts,bm}

\def\eqref#1{equation~\ref{#1}}

\def\1{\bm{1}}

\DeclareMathAlphabet{\mathsfit}{\encodingdefault}{\sfdefault}{m}{sl}
\SetMathAlphabet{\mathsfit}{bold}{\encodingdefault}{\sfdefault}{bx}{n}

\usepackage{booktabs}
\usepackage[table]{xcolor}
\usepackage{makecell}
\usepackage{xcolor}
\usepackage{marvosym}
\usepackage{graphicx} 
\usepackage{enumitem}
\usepackage{bm}
\usepackage{multirow}
\usepackage{colortbl}
\usepackage{wrapfig}
\usepackage{afterpage}
\definecolor{linkblue}{RGB}{0,76,130}
\definecolor{refred}{RGB}{255,0,0}
\definecolor{citeteal}{RGB}{0,128,128}

\newcommand{\appxtosec}[2]{
    \noindent
    \hyperref[#1]{\textcolor[HTML]{3F5F8F}{\textbf{#2}}}
    \par\vspace{0.28em}
}

\newcommand{\appxtosubsec}[2]{
    \noindent\hspace*{1.2em}
    \hyperref[#1]{\textcolor[HTML]{6F8FB8}{#2}}
    \par\vspace{0.14em}
}

\newcommand{\tabnum}[1]{{\footnotesize $#1$}}
\newcommand{\graynum}[1]{{\footnotesize\color{gray!80}$#1$}}

\usepackage[
  colorlinks=true,
  linkcolor=refred,
  citecolor=citeteal,
  urlcolor=linkblue,
  pdfborder={0 0 0}
]{hyperref}
\usepackage{url}

\title{Reconstructing Is Not Acting: \\ Action-Centric Latent Dynamics Modeling}

\author{
    Dingjie Fu$^{1,\diamond}$, Dianxing Shi$^2$, Yangyang Xu$^{1~\textrm{\Letter}}$, Jun Yu$^1$\\
    $^{1}$ Harbin Institute of Technology (Shenzhen)~~~
    $^{2}$ Beihang University \\
    {\tt\small\{dingjiefu1103,cnnlstm\}@gmail.com} \qquad 
}

\begin{document}

\maketitle
\let\thefootnote\relax\footnotetext{$^\diamond$ This work was done when Dingjie Fu was interning at HIT (Shenzhen). ${^{~\textrm{\Letter}}}$ Corresponding author}
\begin{abstract}
Latent action models (LAMs) learn action representations from unlabeled videos by inferring latent actions from visual transitions and reconstructing future states. However, we identify a fundamental \textbf{reconstruction-action mismatch}: lower reconstruction error does not necessarily yield better latent dynamics or downstream performance. We attribute this mismatch to two underconstrained aspects of reconstruction-based latent dynamics modeling: (i) the inverse dynamics model (IDM) is not explicitly encouraged to distinguish action-related transitions from nuisance appearance, and (ii) the forward dynamics model (FDM) can underutilize the inferred latent action by exploiting predictive shortcuts from the current state. To address both limitations, we propose \textbf{ACT-LAM}, a lightweight action-centric framework that strengthens both action extraction and action utilization. Specifically, its Action Query IDM (AQ-IDM) employs learnable action queries and gated aggregation to selectively extract rich action-related transition cues without strong information bottlenecks. And its Action Token FDM (AT-FDM) projects latent actions into action tokens that progressively interact with evolving state representations, enabling continuous state-aware action conditioning. ACT-LAM further streamlines feature processing to concentrate model capacity on latent dynamics modeling. Extensive experiments on several robotic datasets and the VP$^2$ benchmark demonstrate stronger latent action consistency, forward dynamics, and downstream visual planning performance with fewer trainable parameters and lower computational overhead. In particular, ACT-LAM surpasses the previous state of the art by \textbf{7.6\%} on the aggregated VP$^2$ success rate results. Codes at \href{https://github.com/DingjieFu/ACT-LAM}{url}.
\end{abstract}

\section{Introduction}
\label{sec:intro}
Latent action models (LAMs) \citep{iclr24_LAPO,icml24_Genie} learn action representations from videos without action annotations. These representations provide scalable supervision for pretraining vision-language-action (VLA) models \citep{arxiv22_RT1,corl24_OpenVLA,corl25_pi0.5}, reducing their reliance on costly robot action data. A typical LAM consists of two key components, \emph{i.e.}, an inverse dynamics model (IDM) and a forward dynamics model (FDM) \citep{iclr24_LAPO,icml25_LAOM,neurips25_LinearLAM}. Given a pair of consecutive video frames, the IDM infers a latent action that explains their transition, while the FDM reconstructs or predicts the future state conditioned on the current state and the inferred action \citep{iclr25_LAPA,arxiv25_CLAM,icml26_LARA}. Since ground-truth actions are unavailable, reconstruction naturally becomes the primary objective for latent dynamics modeling.

\begin{figure*}[t]
  \centering
   \includegraphics[width=1.0\linewidth]{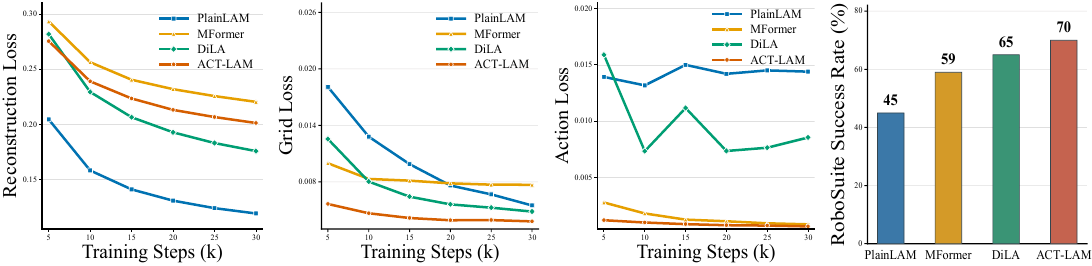}
   \caption{\textbf{Reconstructing is not acting.} We present the validation reconstruction loss, state prediction (grid) loss, and latent action consistency (action) loss over training, followed by the downstream visual planning success rate. Although \emph{PlainLAM} achieves the lowest reconstruction loss, \textbf{ACT-LAM} yields substantially lower grid and action losses and the highest success rate. These results illustrate that the reconstruction quality is not a reliable indicator of latent dynamics modeling or visual planning. The detailed model descriptions are placed in Appendix~\ref{sec:appx:models}.}
   \label{fig:motivation}
\end{figure*}

However, our systematic analysis reveals a significant mismatch: \textbf{lower reconstruction error does not necessarily imply better latent actions.} As illustrated in Fig.~\ref{fig:motivation}, across different LAMs, improvements in future state reconstruction do not consistently translate into better latent action consistency, more accurate forward dynamics, or superior downstream visual planning performance. This discrepancy prompts us to 
consider \emph{why a LAM can reconstruct future states well without learning equally effective latent actions?}

We attribute the mismatch to two underconstrained degrees of freedom in reconstruction-based latent dynamics modeling \citep{arxiv26_41lams}: (i) reconstruction does not uniquely determine what the IDM should extract. This objective cannot distinguish action-related transition cues from nuisance appearance, allowing the IDM to encode both into the latent representation. (ii) reconstruction places no explicit constraint on how strongly the FDM relies on the inferred action. A FDM may exploit predictive shortcuts from the current state and largely bypass the latent action, yielding accurate reconstruction while leaving the inferred action underutilized. Therefore, reconstruction alone is insufficient to ensure both \emph{\textbf{action extraction}} in the IDM and \emph{\textbf{action utilization}} in the FDM. 

Existing LAMs address these two underconstrained aspects unevenly. On the IDM side (\emph{action extraction}), they usually impose strong information bottlenecks. Some methods restrict the capacity of latent actions through vector quantization or explicit regularization \citep{icml24_Genie,icml25_AdaWorld}, whereas others compress rich visual features into compact structure embeddings before action extraction \citep{icml26_DiLA}. While effective, these bottlenecks enforce action relevance only indirectly, prioritizing information restriction over the identification of action-related spatial transition cues. Consequently, they may inadvertently discard action cues entangled with nuisance appearance. By contrast, \emph{action utilization} in the FDM has received considerably less attention. Prior work conditions forward dynamics on latent actions through addition or concatenation \citep{cvpr26_CoMo,icml25_AdaWorld}, cross-attention \citep{iclr25_LAPA, icml26_FLAM}, and AdaLN-zero \citep{icml26_DiLA}. Despite these different mechanisms, the action conditioning generally remains fixed while the state representation evolves, limiting its state-aware interaction throughout forward dynamics. Together, these limitations motivate us to develop a more action-centric framework.

To address the aforementioned limitations, we propose \textbf{ACT-LAM}, an action-centric framework that explicitly strengthens action extraction and action utilization, as shown in Fig.~\ref{fig:framework}. To boost action extraction, we introduce an Action Query IDM (AQ-IDM), which harnesses learnable action queries to directly capture rich spatial transition cues. A gated aggregation then integrates these cues into a continuous latent action, enabling selective action extraction without strong information bottlenecks. To further improve action utilization, we present an Action Token FDM (AT-FDM). It projects the inferred latent action into action tokens and progressively updates them through interaction with evolving state representations. This mechanism enables the continuous state-aware action conditioning in forward dynamics. These complementary designs enable more effective latent dynamics modeling. Meanwhile, ACT-LAM adopts a streamlined architecture, yielding a lightweight model with fewer trainable parameters and lower computational overhead. We conduct extensive experiments on several robotic datasets and the VP$^2$ benchmark, where ACT-LAM consistently improves latent action modeling and downstream visual planning performance. In particular, ACT-LAM surpasses the previous state of the art by \textbf{7.6\%} on the aggregated VP$^2$ success rate, demonstrating the effectiveness of its action-centric design.

In summary, our contributions are as follows:
\begin{itemize}[leftmargin=15pt]
\item We identify the reconstruction-action mismatch in existing LAMs, where lower reconstruction error does not necessarily yield better latent actions, and attribute it to underconstrained action extraction in the IDM and action utilization in the FDM.

\item We propose ACT-LAM, an action-centric framework with AQ-IDM for selectively extracting action-related transition cues and AT-FDM for progressive state-aware action conditioning.

\item Extensive experiments demonstrate that ACT-LAM improves latent action consistency, forward dynamics, and downstream visual planning performance, while using fewer trainable parameters and lower computational overhead.

\end{itemize}

\section{Related Work}
\label{sec:related}
\noindent{\textbf{Learning from human videos.}} 
Leveraging scalable human videos is a promising paradigm for improving real-world robot policy learning \citep{iros21_LbW,corl23_VideoDex,arxiv26_Egoscale}. To reduce the domain gap, some methods use in-domain human demonstrations collected through direct interaction with robot task environments \citep{corl23_MimicPlay,icra25_Egomimic,corl25_HumanPlus}, but such data remain costly to collect at scale. Other works exploit diverse in-the-wild videos to learn general visual representations \citep{corl23_R3M,corl23_MVP} or action priors \citep{corl23_VideoDex}. However, R3M \citep{corl23_R3M} and MVP \citep{corl23_MVP} do not provide explicit action representations, while VideoDex \citep{corl23_VideoDex} relies on predefined motion estimators and specific re-targeting. In this context, latent action models \citep{iclr24_LAPO} offer a pathway towards inferring latent actions from visual transitions, without requiring action annotations or predefined motion cues.

\noindent{\textbf{Latent action models.}} 
LAMs learn action representations from unlabeled videos, which will be used as intermediate supervision for VLA pretraining \citep{iclr25_LAPA,iccv25_Moto,neurips25_LinearLAM} or skill learning \citep{corl23_Xskill,corl25_UniSkill}. Most prior work emphasizes \emph{action extraction} in the IDM, typically by imposing information bottlenecks. Some methods restrict the capacity of latent actions through vector quantization or explicit regularization \citep{iclr24_LAPO,icml24_Genie,arxiv26_41lams}, whereas others compress rich visual features into compact structure embeddings before action extraction \citep{icml26_DiLA}. These approaches encourage compact action representations, but may discard action cues entangled with appearance. Comparatively less attention has been paid to action utilization in the FDM. Prior arts condition forward dynamics on fixed action conditions \citep{icml25_AdaWorld,cvpr26_CoMo,icml26_DiLA}, thereby limiting the state-aware interaction throughout this progress. CoMo \citep{cvpr26_CoMo} is the most similar work to our paper, as both extract continuous latent actions from rich visual representations using learnable queries \citep{icml23_BLIP2}. However, the two approaches differ substantially in action extraction and action utilization. Compared with CoMo, AQ-IDM replaces global self-attention over all tokens with lightweight action-to-patch attention and employs a aggregation strategy for richer action extraction. Moreover, while CoMo uses fixed latent-action conditioning in forward dynamics, AT-FDM progressively updates action tokens through state interaction. These difference enable ACT-LAM to strengthens latent dynamics modeling while retaining a lightweight architecture.

\noindent{\textbf{Learnable Queries.}}
Learnable queries have been widely used to selectively extract task-relevant information from dense representations. Set Transformer \citep{icml19_SetTransformer} introduces learnable vectors for attention-based feature aggregation, while DETR \citep{eccv20_DETR} uses object queries to extract object-specific information from image features. Besides that, these query-based approaches have also been adopted in vision-language models. Flamingo \citep{neurips22_Flamingo} employs a resampler to compress visual features into a fixed number of tokens, and BLIP-2 \citep{icml23_BLIP2} uses learnable queries to extract language-relevant information from visual representations. Motivated by these works, our AQ-IDM introduces action queries for latent dynamics modeling, selectively aggregating action-related spatial transition cues from rich visual representations.

\section{Method}
\label{sec:method}
\begin{figure*}[t]
  \centering
   \includegraphics[width=1.0\linewidth]{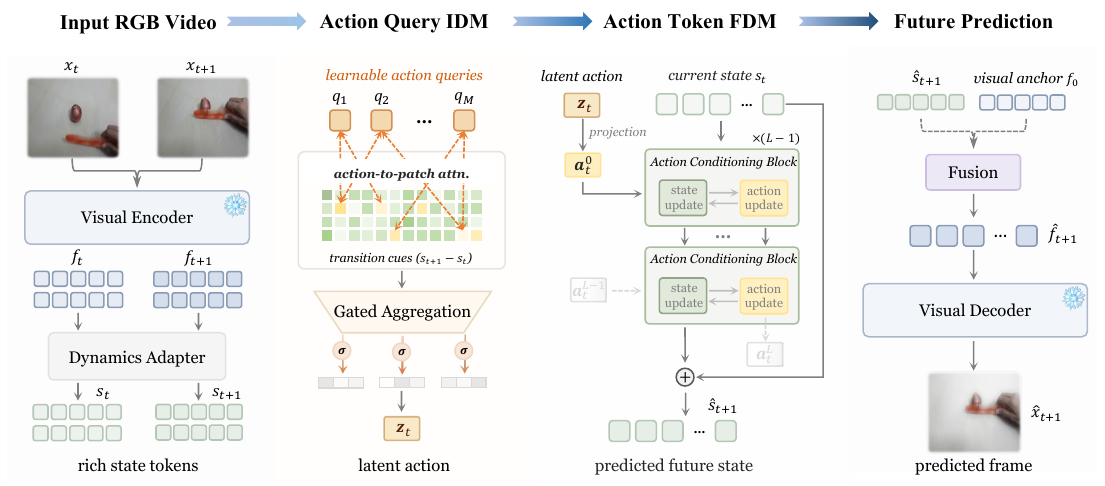}
   \caption{\textbf{The overview of ACT-LAM.} ACT-LAM is a lightweight action-centric framework for latent dynamics modeling. Given consecutive video frames, a frozen visual encoder extracts rich visual features, while the dynamics adapter further encodes them to state tokens. The Action Query IDM deploys learnable action queries to selectively aggregate action-related transition cues into a continuous latent action, which avoids the utilization of strong information bottlenecks. The Action Token FDM projects the inferred latent action into action tokens that are progressively updated through interaction with evolving state representations, enabling state-aware action conditioning. Finally, the predicted future state is decoded into the future frame, fused with with a visual anchor. Our ACT-LAM shows superior performance on downstream visual planning tasks.}
   \label{fig:framework}
\end{figure*}

We introduce \textbf{ACT-LAM}, an action-centric framework designed to address the two underconstrained aspects of latent dynamics modeling: \emph{action extraction} and \emph{action utilization}. As illustrated in Fig.~\ref{fig:framework}, ACT-LAM consists of an Action Query IDM (AQ-IDM) and an Action Token FDM (AT-FDM). AQ-IDM extracts action-related transition cues from rich visual representations, while AT-FDM enables state-aware interaction between action tokens and evolving state representations. The overall architecture is further streamlined to concentrate model capacity on latent dynamics modeling. In this section, we first present the problem formulation and motivate our design in Sec.~\ref{sec:formu}. We then introduce the lightweight action-centric framework in Sec.~\ref{sec:framework}. Finally, we detail the training objectives in Sec.~\ref{sec:overall}.

\subsection{Problem Formulation}
\label{sec:formu}
Let $X=\{x_i\}_{i=1}^{T}$ denote an unlabeled video sequence of $T$ frames. Given a pair of consecutive frames $(x_t, x_{t+1})$, a visual encoder $\mathcal{E}$ extracts their visual embeddings, $(\bm{f}_t, \bm{f}_{t+1})$. Assume that $\bm{s}_t$ denotes the dynamics state of $\bm{f}_t$ and $\bm{s}_t = \mathcal{A}(\bm{f}_t)$, where $\mathcal{A}$ represents a dynamics adapter. A latent action model consists of an inverse dynamics model $\mathcal{I}_{\phi}$ and a forward dynamics model $\mathcal{F}_{\theta}$. The IDM infers a latent action from transitions:
\begin{equation}
\label{eq:idm}
    \bm{z}_t = \mathcal{I}_{\phi}(\bm{s}_t, \bm{s}_{t+1}),
\end{equation}
where the FDM predicts the residual state transition conditioned on latent action $\bm{z}_t$, and the future dynamics state is formulate as:

\begin{equation}
\label{eq:fdm}
    \hat{\bm{s}}_{t+1} =\bm{s}_t+ \Delta\hat{\bm{s}}_t, \quad\text{where} \ \Delta\hat{\bm{s}}_t = \mathcal{F}_{\theta}(\bm{s}_t,\bm{z}_t)
\end{equation}

The predicted dynamics state is then mapped back to the visual feature $\hat{\bm{f}}_{t+1}$ by the feature reconstruction mapping. Without action annotations, the IDM and FDM are jointly optimized by reconstructing the visual feature:
\begin{equation}
\label{eq:rec}
    \mathcal{L}_{\mathrm{rec}} = \|  \bm{f}_{t+1}-\hat{\bm{f}}_{t+1} \|_2.
\end{equation}

However, Eq.~(\ref{eq:rec}) does not explicitly constrain how latent dynamics are learned. In particular, it neither specifies which transition cues should be extracted into $\bm{z}_t$ nor how the FDM utilize latent action during state prediction. Thus, we design ACT-LAM to directly address these two problems.

\subsection{Action-Centric Framework}
\label{sec:framework}
As shown in Fig.~\ref{fig:framework}, ACT-LAM performs latent dynamics modeling in the latent space \citep{arXiv25_VJEPA2,icml26_DiLA}, eliminating the need for training a pixel-space world model \citep{icml25_AdaWorld}. It leverages a frozen visual encoder \citep{tmlr24_DINOv2} to extract rich visual embeddings and an visual decoder \citep{iclr26_RAE} for visual reconstruction. Rather than allocating substantial capacity to feature processing \citep{icml26_DiLA}, ACT-LAM concentrates more on inverse and forward dynamics through AQ-IDM and AT-FDM, respectively.

\noindent{\textbf{Action Query IDM.}}
Existing approaches often promote \emph{action extraction} by imposing strong information bottlenecks, which may discard action-related cues with nuisance appearance. Instead, AQ-IDM preserves rich spatial embeddings and uses learnable action queries as selective readouts to identify action-related transition cues. In this way, action relevance is induced through aggregation rather than aggressive information compression.

As shown in Fig.~\ref{fig:framework}, given a dynamics state sequence $\{\bm{s}_t\}_{t=1}^{T}$, the spatial transition at time $t$ is defined as $\Delta \bm{s}_t = \bm{s}_{t+1}- \bm{s}_t$. A lightweight transition encoder then maps $\Delta \bm{s}_t$ into spatial patch tokens $\bm{E}_t = \{\bm{e}_t\} \in \mathbb{R}^{P \times d_e}$, where $P$ is the number of spatial patch tokens. We add a learnable positional embedding to preserve the spatial information of each token. We introduce $M$ learnable action queries $\bm{Q} = \{\bm{q}_m\} \in \mathbb{R}^{M \times d_q}$ as action-centric readouts \citep{cvpr26_CoMo}. Instead of globally mixing all tokens, each query attends to the spatial transition tokens and computes the action-to-patch attention, formulated as:

\begin{equation}
\begin{aligned}
\bm{A}_t = \operatorname{Softmax}&
((\bm{Q}\bm{W}_q)(\bm{E}_t\bm{W}_k)^{\mathsf T}/{\sqrt{d_k}}), \\
\bm{H}_t &= \bm{A}_t(\bm{E}_t\bm{W}_v),
\end{aligned}
\end{equation}

where $\bm{A}_t\in\mathbb{R}^{M\times P}$ denotes the action-to-patch attention map, and $\bm{H}_t$ represents the resulting query-wise action cues. These cues are subsequently processed by a query mixer and a temporal mixer, followed by gated aggregation to form the latent action: 

\begin{equation}
\bm{\alpha}_t = \operatorname{Softmax}(g(\tilde{\bm{H}}_t)), \quad\bm{C}_t = f_{\mathrm{query}} (\tilde{\bm{H}}_t),
\end{equation}
\begin{equation}
\bm{z}_t = f_{\mathrm{action}}(\bm{\alpha}_t^{\mathsf T}\bm{C}_t),
\end{equation}

where $\bm{\alpha}_t$ and $\bm{C}_t$ represent the importance weights and query-specific action representations, respectively. The gated aggregation selectively combines transition cues across queries, after which $f_{\mathrm{action}}(\cdot)$ maps the aggregated representation to the continuous latent action $\bm{z}_t\in \mathbb{R}^{d_z}$. This allows different queries to capture complementary spatial transition cues while preserving rich information until the aggregation stage. In addition, no vector quantization or explicit capacity regularization \citep{icml25_AdaWorld,icml26_DiLA} is required in AQ-IDM. Moreover, unlike Motion Qformer \citep{cvpr26_CoMo} that applies global self-attention to all tokens, AQ-IDM camputes only action-to-patch attention. This design reduces the attention map from $(2N+1+M)^2$ to $MN$, while enabling each high-capacity query to capture aggregate spatial transition cues.

\noindent{\textbf{Action Token FDM.}}
A key challenge in forward dynamics is to maintain effective action conditioning as the state representation evolves. Prior FDMs typically condition on a fixed latent action \citep{cvpr26_CoMo,icml26_DiLA}, which limits their interaction with states. AT-FDM addresses this limitation by projecting the fixed latent action into an action token that is progressively updated with state representations. Therefore, our design facilitates \emph{action utilization} by state-aware interaction throughout forward dynamics.

At time step $t$, given dynamics state $\bm{s}_t$ and the inferred latent action $\bm{z}_t$, the FDM predicts the future state $\hat{\bm{s}}_{t+1}$ by Eq.~(\ref{eq:fdm}). In AT-FDM, we initialize the state and latent action as $\bm{s}_t^{0} = \bm{s}_t$ and $\bm{a}_t^{0} = f_{\mathrm{proj}}(\bm{z}_t)$, where $\bm{a}_t^{0} \in \mathbb{R}^{d_e}$ is the initial action token. Then, AT-FDM employs a stack of $L$ \emph{Action Conditioning Blocks} (ACBs) to jointly update the state and action representations. At the $l$-th block, they are first processed by self-attention:
\begin{equation}
[\tilde{\bm{s}}_t^{l},\tilde{\bm{a}}_t^{l}] = \operatorname{SA}([\bm{s}_t^{l},\bm{a}_t^{l}]).
\end{equation}

Through this interaction, the action token incorporates the current state context and is subsequently utilized to modulate the state representations \citep{aaai18_FiLM}, defined as:
\begin{equation}
\begin{aligned}
\label{eq:film}
\gamma_t^{l},\beta_t^{l} &= h(\tilde{\bm{a}}_t^{l}), \\
\bm{s}_t^{l+1} = \tilde{\bm{s}}_t^{l} + \sigma(g^{l})
[\gamma_t^{l} &\odot \operatorname{LN} (\tilde{\bm{s}}_t^{l})
+ \beta_t^{l}]
+ \operatorname{FFN}_{s}
(\tilde{\bm{s}}_t^{l}),
\end{aligned}
\end{equation}
where $h(\cdot)$ is the modulation function, $g^{l}$ is a learnable scalar gate
for the $l$-th block, and $\sigma(\cdot)$ denotes the sigmoid function. The Eq.~(\ref{eq:film}) formulates the effect of action modulation, while action token is updated as follows:
\begin{equation}
\bm{a}_t^{l+1} = \tilde{\bm{a}}_t^{l} + \operatorname{FFN}_{a}(\tilde{\bm{a}}_t^{l}).
\end{equation}

Finally, the resulting state representation is projected to predict the residual $\Delta\hat{\bm{s}}_t$. Thus, AT-FDM enables the action token to co-evolve with state representations, thereby ensuring persistent action conditioning throughout forward dynamics.

\begin{wraptable}{r}{0.35\linewidth}
\vspace{-1.0\baselineskip}
\caption{The parameters of major components in LAMs.}
\vspace{1pt}
\begin{tabular}{lcc}
\toprule
Part & DiLA & \textbf{ACT-LAM} \\
\midrule
Proc. & \tabnum{51}\small{M} & \tabnum{1}\small{M} \\
IDM   & \tabnum{36}\small{M} & \tabnum{12}\small{M} \\
FDM   & \tabnum{8}\small{M}  & \tabnum{18}\small{M} \\
Dec.  & \tabnum{28}\small{M} & \tabnum{23}\small{M} \\
\midrule
Total & \tabnum{123}\small{M} & \bm{\tabnum{55}}\small{\textbf{M}} \\
\bottomrule
\end{tabular}
\label{tab:complex}
\end{wraptable}

\noindent{\textbf{Lightweight Designs.}}
ACT-LAM further streamlines feature processing to concentrate model capacity on latent dynamics modeling. As summarized in Tab.~\ref{tab:complex}, a considerable portion of the trainable parameters in DiLA \citep{icml26_DiLA} is devoted to feature processing (\emph{i.e.}, ST-Transformer) rather than to the IDM and FDM themselves. Contrarily, ACT-LAM adopts a lightweight pipeline together with the proposed AQ-IDM and AT-FDM. Specifically, we remove the ST-Transformer \citep{arXiv20_STT}, replace the DeepConv IDM with our AQ-IDM, and adopt our AT-FDM to strengthen action conditioning. Empirical results demonstrate the efficiency and effectiveness of these designs. Details are shown in Appendix~\ref{sec:appx:config} and \ref{sec:appx:params}. 

\subsection{Training Objectives}
\label{sec:overall}
Following DiLA \citep{icml26_DiLA}, we adopt its teacher-forcing training paradigm while adapting the loss formulation to ACT-LAM. The resulting objectives are defined as follows:
\begin{equation}
\begin{aligned}
\label{eq:losses}
    \mathcal{L}_{\mathrm{rec}} &= \| \bm{f}_{t} - \hat{\bm{f}}_{t} \|_2, \\
    \mathcal{L}_{\mathrm{grid}} &= \operatorname{SmoothL1}(\Delta \bm{s}_{t} - \mathcal{F}_{\theta}(\bm{s}_t,\bm{z}_t)), \\
    \mathcal{L}_{\mathrm{act}} &= \operatorname{SmoothL1}[\mathcal{I}_{\phi}(\bm{s}_{t+1}-\bm{s}_{t}) - \mathcal{I}_{\phi}(\hat{\bm{s}}_{t+1}-\bm{s}_{t})], \\
    \mathcal{L}_{\mathrm{sym}} &= \| \bm{z}_{t}\|_2 + \frac{\sum_{t} \cdot (\cos(\bm{z}_t^{\mathrm{fwd}}, \bm{z}_t^{\mathrm{bwd}})+1)}{T-1}, \\
    \mathcal{L}_{\mathrm{var}} &= \sum\nolimits_{d}(\sigma - \operatorname{Std}(\bm{z}_t)) / d_z, 
\end{aligned}
\end{equation}
where $\mathcal{L}_{\mathrm{rec.}}$, $\mathcal{L}_{\mathrm{grid}}$, and $\mathcal{L}_{\mathrm{act.}}$ serve as indicators of reconstruction quality, forward state prediction, and latent action consistency, respectively. We retain $\mathcal{L}_{\mathrm{sym.}}$ to encourage the forward and backward symmetry, and introduce $\mathcal{L}_{\mathrm{var.}}$ to promote latent action diversity. In general, the total training objective is a weighted combination, defined as:
\begin{equation}
\label{eq:all_losses}
    \mathcal{L}_{all} = \lambda_{r}\mathcal{L}_{\mathrm{rec}} + \lambda_{g}\mathcal{L}_{\mathrm{grid}} + \lambda_{a}\mathcal{L}_{\mathrm{act}} + \lambda_{s}\mathcal{L}_{\mathrm{sym}} + \lambda_{v}\mathcal{L}_{\mathrm{var}}.
\end{equation}

\section{Experiments}
\label{sec:exper}
In this section, we empirically evaluate ACT-LAM to assess its effectiveness and key properties. We first introduce the experimental setup in Sec.~\ref{sec:setups}, including implementations and datasets. We then study ACT-LAM along the two key aspects motivated in Sec.~\ref{sec:method}: \emph{action extraction} and \emph{action utilization}. We examine whether AQ-IDM learns more effective latent actions and whether AT-FDM better exploits them for forward dynamics modeling in Secs.~\ref{sec:ae} and~\ref{sec:au}, respectively. Subsequently, we assess the effectiveness of our designs in downstream visual planning in Sec.~\ref{sec:vp}. Finally, we provide additional analyses on the reconstruction-action mismatch and model efficiency in Sec.~\ref{sec:analysis}.

\subsection{Experimental Setup}
\label{sec:setups}
\noindent{\textbf{Implementation Details.}}
We adopt a frozen DINOv2 \citep{tmlr24_DINOv2} as the visual encoder and use the ViT-XL RAE decoder \citep{iclr26_RAE} for visual reconstruction. ACT-LAM is trained only in the teacher-forcing paradigm. We optimize all trainable components using AdamW \citep{iclr17_AdamW} with a learning rate of $1 \times 10^{-4}$. Unless otherwise specified, all methods are trained using the same visual backbone, training data, and optimizer for fair comparison. More architectural and training details are given in Appendix~\ref{sec:appx:details}.

\noindent{\textbf{Benchmark Datasets.}}
Following DiLA, we employ four video datasets to pretrain the latent action models: Something-Something-v2 (SSv2) \citep{iccv17_SSv2}, RT-1 \citep{arxiv22_RT1}, RECON \citep{corl22_RECON}, and LoopNav \citep{arxiv25_LoopNav}. We then evaluate the pretrained models on the $\text{VP}^2$ benchmark \citep{iclr23_VP2} to assess their downstream visual planning performance. Refer to Appendix~\ref{sec:appx:datasets} for detailed descriptions.

\begin{table*}[t]
\centering
\caption{\textbf{Linear probing MSE across two datasets.} Results represent the mean and standard deviation of 4 independent runs per task. The best results are highlighted in ``\textbf{bold}".}
\vspace{0pt}
\label{tab:lp}
\begin{center}
\resizebox{0.7\textwidth}{!}{
\begin{tabular}{c c c c >{\columncolor{purple!3}}c}
    \toprule
    \textbf{Dataset} & AdaWorld & MFormer & DiLA & ACT-LAM \\
    \midrule
    {Push-T} 
    & \tabnum{0.033\pm0.001}
    & \tabnum{0.016\pm0.001} 
    & \tabnum{0.009\pm0.003}
    & \bm{\tabnum{0.007\pm0.000}} \\
    
    {Block Pushing} 
    & \tabnum{0.039\pm0.005} 
    & \tabnum{0.048\pm0.003}
    & \tabnum{0.037\pm0.013}
    & \bm{\tabnum{0.031\pm0.001}} \\
    \bottomrule
    \vspace{-2.5em}
\end{tabular}}
\end{center}
\end{table*}

\begin{figure}[t]
    \centering
    \begin{minipage}[t]{0.48\linewidth}
        \centering
        \includegraphics[width=\linewidth]{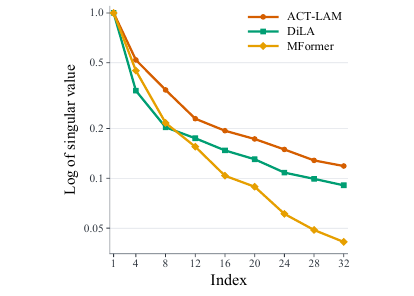}
        \vspace{-2.2em}
        \caption{\textbf{Latent action singular-value spectra on Block Pushing.} ACT-LAM exhibits a slower spectral decay, indicating a richer and more diverse latent action representation.}
        \label{fig:singular}
    \end{minipage}
    \hfill
    \begin{minipage}[t]{0.48\linewidth}
        \centering
        \includegraphics[width=\linewidth]{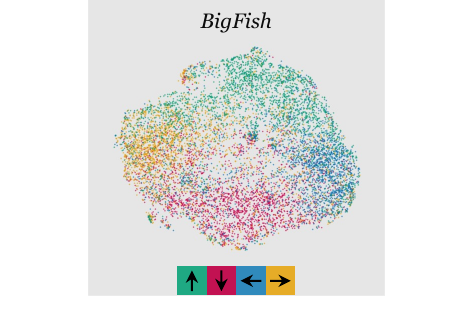}
        \vspace{-2.2em}
        \caption{\textbf{UMAP visualization on BigFish.} Each point represents a latent action extracted by the pretrained AQ-IDM. Arrows in the legend correspond to movement directions.}
        \label{fig:umap}
    \vspace{0.5em}
    \end{minipage}
    \begin{minipage}[t]{0.98\linewidth}
        \centering
        \includegraphics[width=\linewidth]{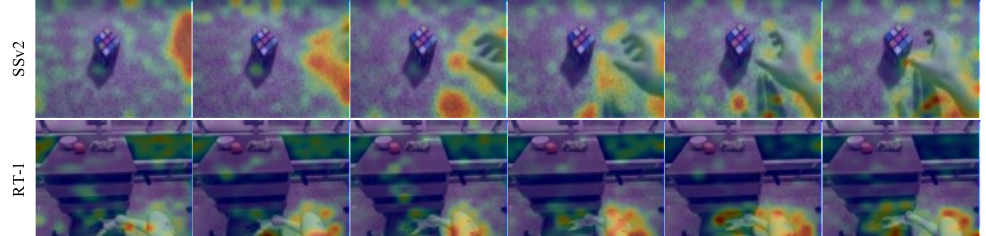}
        \vspace{-2em}
        \caption{\textbf{Action-to-patch attention heatmaps for two samples.} We visualize one video clip from each of the SSV2 and RT-1 datasets. The learned queries primarily attend to action-related regions, indicating their ability to extract action-related spatial transition cues.}
        \label{fig:a2p}
    \end{minipage}
    \vspace{-1.5em}
\end{figure}

\subsection{Action Extraction}
\label{sec:ae}
\noindent{\textbf{Linear Probing.}} 
To evaluate the quality of the learned latent action representations, we perform linear probing on two benchmarks unseen during pretraining: Block Pushing \citep{corl22_BlockPushing} and Push-T \citep{ijrr25_PushT}. As shown in Tab.~\ref{tab:lp}, ACT-LAM achieves the lowest probing Mean Squared Error (MSE) on the two datasets, indicating that its latent actions are more linearly aligned with the robot actions and preserve more action-relevant information. It is noteworthy that the reported AdaWorld \citep{icml25_AdaWorld} results are obtained using the official pretrained weights released by the authors. More linear probing results and ablations are given in Appendix~\ref{sec:appx:lp}.

\noindent{\textbf{Action Diversity.}} 
As shown in Fig.~\ref{fig:singular}, we report the normalized logarithm of singular values \citep{icml23_RankMe} of latent actions on the Block Pushing dataset. Compared with other methods, ACT-LAM exhibits a slower singular-value decay, indicating a higher effective dimensionality and more diverse latent action representations. Noteworthy, this increased diversity is accompanied by the lowest MSE error in Tab.~\ref{tab:lp}, suggesting that ACT-LAM captures richer variations while preserving strong alignment with robot actions. These results demonstrate that AQ-IDM extracts more informative and action-relevant transition cues. Details are placed in Appendix~\ref{sec:appx:diversity}.

\noindent{\textbf{UMAP Visualization.}} 
To provide a more intuitive illustration, we utilize UMAP \citep{arxiv18_UMAP} to visualize latent actions on the \emph{BigFish} environment from the Procgen environment \citep{icml18_Procgen}. As shown in Fig.~\ref{fig:umap}, AQ-IDM forms distinct clusters that correspond to the true direction actions. More visualizations are detailed in Appendix ~\ref{sec:appx:umaps}.

\noindent{\textbf{Action-to-Patch Attention.}} 
To further demonstrate the effectiveness of the learnable action queries, we visualize the action-to-patch attention heatmaps \citep{cvpr16_CAM}. As shown in Fig.~\ref{fig:a2p}, the action queries consistently focus on regions associated with salient motion in both SSV2 and RT-1. These attention patterns indicate that AQ-IDM can selectively aggregate action-related transition cues from rich visual representations, providing evidence for its action-centric extraction approach. Additional visualizations and per-query attention maps are provided in Appendix~\ref{sec:appx:a2p}.

\begin{table*}[t]
\centering
\begin{minipage}[t]{0.48\textwidth}
\vspace{0pt}
\centering
\caption{\textbf{Rollout feature reconstruction MSE on SSv2 and RT-1.}
@$k$ denotes the MSE averaged over the first $k$ predicted steps.
Results are reported as mean $\pm$ standard deviation.}
\label{tab:rollout_mse}
\vspace{0.3em}
\setlength{\tabcolsep}{5pt}
\renewcommand{\arraystretch}{1.4}
\resizebox{\linewidth}{!}{
\begin{tabular}{cccccc}
\toprule
\multirow{2}{*}{\textbf{Dataset}}
& \multirow{2}{*}{\textbf{Model}}
& \multicolumn{4}{c}{\textbf{Recon. MSE} $\bm{\downarrow}$} \\
\cmidrule(lr){3-6}
& & @1 & @4 & @8 & @15 \\
\midrule
\multirow{2}{*}{SSv2}
& DiLA 
& \tabnum{0.355\pm0.118}
& \tabnum{0.410\pm0.113} 
& \tabnum{0.452\pm0.106} 
& \bm{\tabnum{0.495\pm0.100}} \\
& \cellcolor{purple!3}ACT-LAM 
& \cellcolor{purple!3}\bm{\tabnum{0.175\pm0.084}}
& \cellcolor{purple!3}\bm{\tabnum{0.266\pm0.102}}
& \cellcolor{purple!3}\bm{\tabnum{0.428\pm0.099}}
& \cellcolor{purple!3}\tabnum{0.667\pm0.095} \\
\midrule
\multirow{2}{*}{RT-1}
& DiLA 
& \tabnum{0.215\pm0.069}
& \tabnum{0.251\pm0.070}
& \bm{\tabnum{0.284\pm0.073}}
& \bm{\tabnum{0.323\pm0.078}} \\
& \cellcolor{purple!3}ACT-LAM 
& \cellcolor{purple!3}\bm{\tabnum{0.119\pm0.051}}
& \cellcolor{purple!3}\bm{\tabnum{0.211\pm0.068}}
& \cellcolor{purple!3}\tabnum{0.442\pm0.070}
& \cellcolor{purple!3}\tabnum{0.705\pm0.066} \\
\bottomrule
\end{tabular}}
\end{minipage}
\hfill
\begin{minipage}[t]{0.48\textwidth}
\vspace{0pt}
\centering
\caption{\textbf{Video generation metrics on SSv2 and RT-1.}
We utilize SSIM and LPIPS for evaluation. {\color{gray!80}\textit{Recon.}} denotes the direct encode-decode reconstruction of the input frames.}
\label{tab:ssim}
\vspace{0.3em}
\setlength{\tabcolsep}{4pt}
\renewcommand{\arraystretch}{1.32}
\resizebox{\linewidth}{!}{
\begin{tabular}{ccccc}
\toprule
\multirow{2}{*}{\textbf{Model}}
& \multicolumn{2}{c}{\textbf{SSv2}}
& \multicolumn{2}{c}{\textbf{RT-1}} \\
\cmidrule(r){2-3}
\cmidrule(l){4-5}
& SSIM $\uparrow$
& LPIPS $\downarrow$
& SSIM $\uparrow$
& LPIPS $\downarrow$ \\
\midrule
\color{gray!80} \textit{Recon.}
& \graynum{0.717\pm0.141}
& \graynum{0.125\pm0.038}
& \graynum{0.621\pm0.072}
& \graynum{0.147\pm0.023} \\
AdaWorld (FDM)
& \tabnum{0.557\pm0.160}
& \tabnum{0.723\pm0.130}
& \tabnum{0.377\pm0.074}
& \tabnum{0.713\pm0.090} \\
DiLA
& \bm{\tabnum{0.578\pm0.150}}
& \tabnum{0.637\pm0.130}
& \bm{\tabnum{0.506\pm0.082}}
& \bm{\tabnum{0.364\pm0.125}} \\
\rowcolor{purple!3} ACT-LAM
& \tabnum{0.508\pm0.148}
& \bm{\tabnum{0.430\pm0.087}}
& \tabnum{0.375\pm0.060}
& \tabnum{0.438\pm0.049} \\
\bottomrule
\end{tabular}}
\vspace{0.3em}
\end{minipage}
\end{table*}

\begin{figure}[t]
  \centering
   \includegraphics[width=1.0\linewidth]{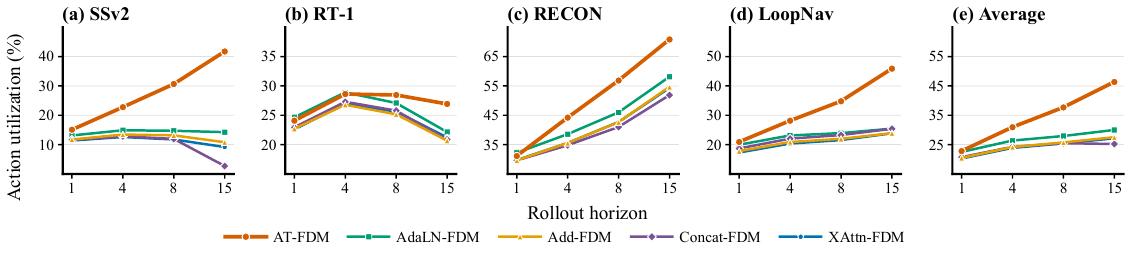}
   \vspace{-2em}
   \caption{\textbf{Action utilization across rollout horizons.} AT-FDM consistently achieves larger utilization gains than other action-conditioning schemes. (Some curves overlap and may be occluded.)}
   \label{fig:action_utilization}
   \vspace{-1.5em}
\end{figure}

\subsection{Action Utilization}
\label{sec:au}
\noindent{\textbf{Rollout MSE.}}
To evaluate forward dynamics under latent action conditioning, we report multi-step rollout MSE in Tab.~\ref{tab:rollout_mse}. ACT-LAM achieves lower errors at short horizons, while DiLA performs better at longer horizons. \emph{Notably, ACT-LAM is trained solely with teacher-forcing, whereas DiLA uses its official rollout-finetuned checkpoint.} Details are provided in Appendix~\ref{sec:appx:rollout}.

\noindent{\textbf{Video Generation.}}
We evaluate the generation fidelity using autoregressive rollouts on SSv2 and RT-1. We report SSIM \citep{tip04_SSIM} to measure structural similarity and LPIPS \citep{cvpr18_LPIPS} to assess perceptual similarity. As shown in Tab.~\ref{tab:ssim}, ACT-LAM achieves the best LPIPS on SSv2, while DiLA performs better on several other reconstruction metrics. These results suggest that ACT-LAM maintains reasonable generation fidelity, although its primary gains lie in learning more effective action-centric dynamics. This observation is also consistent with our main finding that \emph{stronger reconstruction performance does not necessarily translate into better latent action modeling or downstream control}. The detailed descriptions are given in Appendix~\ref{sec:appx:gene}.

\noindent{\textbf{Action Transfer.}}
We further examine whether the inferred latent actions can transfer across visual contexts by applying 16-step action sequences through autoregressive generation. The detailed descriptions and qualitative results are provided in Appendix~\ref{sec:appx:action_transfer}.

\noindent{\textbf{Utilization Analysis.}}
To assess how effectively AT-FDM utilizes latent actions, we introduce an action utilization evaluation. Specifically, we freeze the pretrained IDM and train only different FDMs. We then replace the original latent action sequence with a \emph{constant action} and measure the resulting relative increase in rollout state MSE. As shown in Fig.~\ref{fig:action_utilization}, AT-FDM consistently achieves higher action utilization, particularly at longer horizons, demonstrating its stronger ability to leverage latent actions during forward dynamics modeling. More details are provided in Appendix~\ref{sec:appx:util}.

\subsection{Visual Planning}
\label{sec:vp}
To assess the effectiveness of ACT-LAM in robotic control, we evaluate the pretrained LAMs on the VP$^2$ benchmark \citep{iclr23_VP2}. Following prior works \citep{icml25_AdaWorld,icml26_DiLA}, we first use the pretrained IDM to extract latent actions from state transitions in the downstream datasets. Then, we learn a action-to-latent MLP to map ground-truth actions into the learned latent action space. Subsequently, the MLP is jointly finetuned with the pretrained FDM on robotic trajectories, enabling the resulting dynamics model to perform action-conditioned visual prediction for planning. All experiments follow the evaluation protocol established by AdaWorld \citep{icml25_AdaWorld}.

After adaptation, the pretrained LAM is converted into an action-conditioned world model and utilized as the dynamics model for model predictive control. Following prior works, we adopt the sampling-based Model Predictive Path Integral (MPPI) algorithm \citep{icra16_MPPI} to optimize action sequences according to the predicted future visual states. As reported in Tab.~\ref{tab:vp2}, ACT-LAM surpasses AdaWorld and DiLA on the majority of tasks. In particular, on ``Push Blue Button" and ``Push Green Button", ACT-LAM improves over the previous state of the art by $11.5\%$ and $29.2\%$, respectively. These results further demonstrate that superior latent dynamics modeling does not necessarily require high visual reconstruction quality, as ACT-LAM achieves stronger latent dynamics despite its relatively weaker reconstruction performance. More details are given in Appendix~\ref{sec:appx:vp}.

\begin{table*}[t]
\centering
\caption{\textbf{Visual planning performance in $\text{VP}^2$ benchmark.}
We report the success rates across 5 tasks in RoboDesk and 1 task in RoboSuite, and the aggregated success rate is calculated by normalizing against the Simulator’s success rate. Results represent the mean and standard deviation of the success rates on average across 4 random seeds.}
\vspace{0.3em}
\label{tab:vp2}
\setlength{\tabcolsep}{4pt}
\renewcommand{\arraystretch}{1.3}
\resizebox{0.98\textwidth}{!}{
\begin{tabular}{cccccccc}
    \toprule
    \multirow{2}{*}{\raisebox{-1ex}{\textbf{Method}}}
    & \multicolumn{6}{c}{\textbf{Success Rate (\%)} $\uparrow$}
    & \multirow{2}{*}{\raisebox{-1ex}{\textbf{Aggregate} $\uparrow$}} \\
    \cmidrule(lr){2-7}
    & \shortstack{Robosuite\\Push}
    & \shortstack{Open\\Slide}
    & \shortstack{Blue\\Button}
    & \shortstack{Green\\Button}
    & \shortstack{Red\\Button}
    & \shortstack{Upright\\Block}
    & \\
    \midrule

    \color{gray!80} \small{Simulator}
    & \graynum{89.25^{\pm2.49}}
    & \graynum{64.17^{\pm7.22}}
    & \graynum{100.00^{\pm0.00}}
    & \graynum{89.17^{\pm4.93}}
    & \graynum{93.33^{\pm2.36}}
    & \graynum{95.00^{\pm1.67}}
    & \graynum{100.00} \\
    
    \small{AdaWorld}
    & \tabnum{63.50^{\pm1.71}}
    & \tabnum{5.83^{\pm2.85}}
    & \tabnum{29.17^{\pm2.50}}
    & \tabnum{10.83^{\pm2.50}}
    & \tabnum{10.00^{\pm2.36}}
    & \tabnum{5.00^{\pm0.96}}
    & \tabnum{21.54} \graynum{(22.92)} \\
    
    \small{DiLA}
    & \tabnum{68.00^{\pm1.41}}
    & \tabnum{\bm{15.00^{\pm5.00}}}
    & \tabnum{78.33^{\pm3.73}}
    & \tabnum{35.83^{\pm4.93}}
    & \tabnum{20.83^{\pm5.95}}
    & \tabnum{3.33^{\pm2.72}}
    & \tabnum{41.44} \graynum{(40.65)} \\
    \midrule
    
    \rowcolor{purple!3} \small{ACT-LAM}
    & \tabnum{\bm{71.25^{\pm3.63}}}
    & \tabnum{10.83^{\pm2.76}}
    & \tabnum{\bm{90.83^{\pm2.76}}}
    & \tabnum{\bm{65.00^{\pm7.64}}}
    & \tabnum{\bm{25.00^{\pm3.73}}}
    & \tabnum{\bm{6.67^{\pm2.36}}}
    & \tabnum{\bm{49.04}}\textsuperscript{\textcolor{red}{
    \raisebox{0.05ex}{$\bm{\uparrow}$}\hspace{-0.3em}\textbf{7.6\%}}} \\
    \bottomrule
\end{tabular}}
\vspace{-1.0em}
\end{table*}

\subsection{Further Analyses}
\label{sec:analysis}
ACT-LAM achieves strong performance while maintaining a lightweight and computationally efficient design. Detailed efficiency evaluations, including FLOPs, FPS, and GPU memory usage, are provided in Appendix~\ref{sec:appx:efficiency} due to space constraints. Additional analyses and supplementary experiments are presented in Appendix~\ref{sec:appx:analyses}.

\section{Conclusion}
In this paper, we identify a reconstruction-action mismatch in existing latent action models and introduce ACT-LAM to address this mismatch by strengthening both \emph{action extraction} and \emph{action utilization}. Extensive experiments and analyses demonstrate consistent improvements in latent action quality, forward dynamics, and downstream planning. Notably, ACT-LAM achieves these gains with fewer trainable parameters and lower computational overhead, offering a more effective and efficient approach to latent dynamics modeling.

\noindent{\textbf{Limitations.}}
While ACT-LAM achieves stronger action-centric latent dynamics, its pixel-level reconstruction fidelity remains limited, especially over long autoregressive rollouts. Our evaluation is also restricted to the current benchmarks and does not yet cover settings that demand highly accurate visual prediction or fine-grained control. The results therefore show that reconstruction quality is not a reliable indicator for action quality within the evaluated settings. Broader validation is needed to assess how generally this mismatch holds. Future work could improve visual fidelity while preserving the action-centric properties of the learned dynamics.


\section*{AI use statement}
In this work, we used generative AI tools to assist with code development and language polishing. Specifically, generative AI was used to code implementation and debugging, as well as to improve the grammar of this manuscript. AI-assisted code was manually reviewed and tested by the authors before being used in our experiments. We did not use generative AI tools to autonomously conduct the research or generate the manuscript. In general, we have reviewed all AI-assisted work and take full responsibility for the final content of this paper.

\section*{Reproducibility statement}
We make several efforts to ensure the reproducibility of our results. The architecture and training objectives of ACT-LAM are described in detail in Sec.~\ref{sec:method}. And the datasets, implementation settings, and evaluation protocols are provided in Sec.~\ref{sec:exper} and the corresponding experimental sections. Additional configurations and implementation details are given in Appendix~\ref{sec:appx:details} and \ref{sec:appx:results}. We also report the settings used for the downstream visual planning evaluation and the corresponding ablation studies to facilitate consistent reproduction and comparison.

\subsubsection*{Acknowledgments}
We would like to thank \emph{Tianqiu Zhang} from Peking University for helpful suggestions, and \emph{Yujun Zhang}, \emph{Chenxin Yuan}, and \emph{Lu Pan} for feedback on the draft.

\bibliography{iclr2027_conference}

@string{CVPR={CVPR}}

@string{TIP={IEEE TIP}}

@string{ICCV={ICCV}}

@string{CoRL={CoRL}}

@string{IROS={IROS}}

@string{ICRA={ICRA}}

@string{ECCV={ECCV}}

@string{NeurIPS={NeurIPS}}

@string{ICML={ICML}}

@string{ICLR={ICLR}}

@string{AAAI={AAAI}}

@inproceedings{iclr24_LAPO,
  title={Learning to act without actions},
  author={Schmidt, Dominik and Jiang, Minqi},
  booktitle=ICLR,
  pages={9379--9395},
  year={2024}
}

@inproceedings{icml24_Genie,
  title={Genie: Generative interactive environments},
  author={Bruce, Jake and Dennis, Michael D and Edwards, Ashley and Parker-Holder, Jack and Shi, Yuge and Hughes, Edward and Lai, Matthew and Mavalankar, Aditi and Steigerwald, Richie and Apps, Chris and others},
  booktitle=ICML,
  year={2024}
}

@article{arxiv22_RT1,
  title={Rt-1: Robotics transformer for real-world control at scale},
  author={Brohan, Anthony and Brown, Noah and Carbajal, Justice and Chebotar, Yevgen and Dabis, Joseph and Finn, Chelsea and Gopalakrishnan, Keerthana and Hausman, Karol and Herzog, Alex and Hsu, Jasmine and others},
  journal={arXiv preprint arXiv:2212.06817},
  year={2022}
}

@inproceedings{corl24_OpenVLA,
  title={OpenVLA: An Open-Source Vision-Language-Action Model},
  author={Kim, Moo Jin and Pertsch, Karl and Karamcheti, Siddharth and Xiao, Ted and Balakrishna, Ashwin and Nair, Suraj and Rafailov, Rafael and Foster, Ethan P and Sanketi, Pannag R and Vuong, Quan and others},
  booktitle=CoRL,
  year={2024}
}

@inproceedings{corl25_pi0.5,
  title     = {{$\pi_{0.5}$}: A Vision-Language-Action Model with Open-World Generalization},
  author    = {Black, Kevin and Brown, Noah
               and Darpinian, James and Dhabalia, Karan and Driess, Danny
               and Esmail, Adnan and Equi, Michael and Finn, Chelsea
               and Fusai, Niccolo and others},
  booktitle = CoRL,
  year      = {2025}
}

@article{arxiv25_CLAM,
  title={Clam: Continuous latent action models for robot learning from unlabeled demonstrations},
  author={Liang, Anthony and Czempin, Pavel and Hong, Matthew M and Zhou, Yutai and Wang, Jingzhen and Biyik, Erdem and Tu, Stephen},
  journal={arXiv preprint arXiv:2505.04999},
  year={2025}
}

@inproceedings{iclr25_LAPA,
  title={Latent action pretraining from videos},
  author={Ye, Seonghyeon and Jang, Joel and Jeon, Byeongguk and Joo, Se June and Yang, Jianwei and Peng, Baolin and Mandlekar, Ajay and Tan, Reuben and Chao, Yu-Wei and Lin, Bill Yuchen and others},
  booktitle=ICLR,
  pages={28213--28239},
  year={2025}
}

@inproceedings{icml25_AdaWorld,
  title={AdaWorld: learning adaptable world models with latent actions},
  author={Gao, Shenyuan and Zhou, Siyuan and Du, Yilun and Zhang, Jun and Gan, Chuang},
  booktitle=ICML,
  pages={18744--18771},
  year={2025}
}

@inproceedings{icml25_LAOM,
  title={Latent Action Learning Requires Supervision in the Presence of Distractors},
  author={Nikulin, Alexander and Zisman, Ilya and Tarasov, Denis and Nikita, Lyubaykin and Polubarov, Andrei and Kiselev, Igor and Kurenkov, Vladislav},
  booktitle=ICML,
  pages={46427--46447},
  year={2025},
  organization={PMLR}
}

@inproceedings{neurips25_LinearLAM,
  title={What do latent action models actually learn?},
  author={Zhang, Chuheng and Pearce, Tim and Zhang, Pushi and Wang, Kaixin and Chen, Xiaoyu and Shen, Wei and Zhao, Li and Bian, Jiang},
  booktitle=NeurIPS,
  pages={146676--146697},
  year={2026}
}

@inproceedings{cvpr26_CoMo,
  title={Como: Learning continuous latent motion from internet videos for scalable robot learning},
  author={Yang, Jiange and Shi, Yansong and Zhu, Haoyi and Liu, Mingyu and Ma, Kaijing and Wang, Yating and Wu, Gangshan and He, Tong and Wang, Limin},
  booktitle=CVPR,
  pages={42352--42363},
  year={2026}
}

@inproceedings{icml26_DiLA,
  title={DiLA: Disentangled Latent Action World Models},
  author={Zhang, Tianqiu and Lyu, Muyang and Zhang, Yufan and Fang, Fang and Wu, Si},
  booktitle=ICML,
  year={2026}
}

@inproceedings{icml26_FLAM,
  title={Factored Latent Action World Models},
  author={Wang, Zizhao and Shi, Chang and Hu, Jiaheng and Rohling, Kevin and Mart{\'\i}n-Mart{\'\i}n, Roberto and Zhang, Amy and Stone, Peter},
  booktitle=ICML,
  year={2026}
}

@inproceedings{icml26_LARA,
  title={LARA: Latent Action Representation Alignment for Vision-Language-Action Models},
  author={Liu, Mengya and Jia, Baoxiong and Huang, Jiangyong and Zhang, Jingze and Huang, Siyuan},
  booktitle=ICML,
  year={2026}
}

@inproceedings{iccv25_Moto,
  title={Moto: Latent motion token as the bridging language for robot manipulation},
  author={Chen, Yi and Ge, Yuying and Li, Yizhuo and Ge, Yixiao and Ding, Mingyu and Shan, Ying and Liu, Xihui},
  booktitle=ICCV,
  year={2025}
}

@inproceedings{iros21_LbW,
  title={Learning by watching: Physical imitation of manipulation skills from human videos},
  author={Xiong, Haoyu and Li, Quanzhou and Chen, Yun-Chun and Bharadhwaj, Homanga and Sinha, Samarth and Garg, Animesh},
  booktitle=IROS,
  pages={7827--7834},
  year={2021},
  organization={IEEE}
}

@inproceedings{corl23_VideoDex,
  title={Videodex: Learning dexterity from internet videos},
  author={Shaw, Kenneth and Bahl, Shikhar and Pathak, Deepak},
  booktitle=CoRL,
  pages={654--665},
  year={2023},
  organization={PMLR}
}

@inproceedings{corl23_MimicPlay,
  title={MimicPlay: Long-Horizon Imitation Learning by Watching Human Play},
  author={Wang, Chen and Fan, Linxi and Sun, Jiankai and Zhang, Ruohan and Fei-Fei, Li and Xu, Danfei and Zhu, Yuke and Anandkumar, Anima},
  booktitle=CoRL,
  pages={201--221},
  year={2023},
  organization={PMLR}
}

@inproceedings{icra25_Egomimic,
  title={Egomimic: Scaling imitation learning via egocentric video},
  author={Kareer, Simar and Patel, Dhruv and Punamiya, Ryan and Mathur, Pranay and Cheng, Shuo and Wang, Chen and Hoffman, Judy and Xu, Danfei},
  booktitle=ICRA,
  pages={13226--13233},
  year={2025},
  organization={IEEE}
}

@inproceedings{corl25_HumanPlus,
  title={HumanPlus: Humanoid Shadowing and Imitation from Humans},
  author={Fu, Zipeng and Zhao, Qingqing and Wu, Qi and Wetzstein, Gordon and Finn, Chelsea},
  booktitle=CoRL,
  pages={2828--2844},
  year={2025},
  organization={PMLR}
}

@inproceedings{corl23_R3M,
  title={R3M: A Universal Visual Representation for Robot Manipulation},
  author={Nair, Suraj and Rajeswaran, Aravind and Kumar, Vikash and Finn, Chelsea and Gupta, Abhinav},
  booktitle=CoRL,
  pages={892--909},
  year={2023},
  organization={PMLR}
}

@inproceedings{corl23_MVP,
  title={Real-world robot learning with masked visual pre-training},
  author={Radosavovic, Ilija and Xiao, Tete and James, Stephen and Abbeel, Pieter and Malik, Jitendra and Darrell, Trevor},
  booktitle=CoRL,
  pages={416--426},
  year={2023},
  organization={PMLR}
}

@article{arxiv26_Egoscale,
  title={Egoscale: Scaling dexterous manipulation with diverse egocentric human data},
  author={Zheng, Ruijie and Niu, Dantong and Xie, Yuqi and Wang, Jing and Xu, Mengda and Jiang, Yunfan and Casta{\~n}eda, Fernando and Hu, Fengyuan and Tan, You Liang and Fu, Letian and others},
  journal={arXiv preprint arXiv:2602.16710},
  year={2026}
}

@inproceedings{corl23_Xskill,
  title={Xskill: Cross embodiment skill discovery},
  author={Xu, Mengda and Xu, Zhenjia and Chi, Cheng and Veloso, Manuela and Song, Shuran},
  booktitle=CoRL,
  pages={3536--3555},
  year={2023},
  organization={PMLR}
}

@inproceedings{corl25_UniSkill,
  title={UniSkill: Imitating Human Videos via Cross-Embodiment Skill Representations},
  author={Kim, Hanjung and Kang, Jaehyun and Kang, Hyolim and Cho, Meedeum and Kim, Seon Joo and Lee, Youngwoon},
  booktitle=CoRL,
  year={2025}
}

@article{arxiv26_41lams,
  title={What Matters for Latent Actions in Robot Learning},
  author={Bu, Xizhou and Hu, Qingda and Zhou, Lei and Zhang, Lingfeng and Tang, Yingbo and Liu, Zihao and Tao, Xinyi and Ma, Zhiqiang and Huang, Qingqiu and Tang, Chufeng and others},
  journal={arXiv preprint arXiv:2608.19613},
  year={2026}
}

@article{tmlr24_DINOv2,
  title={DINOv2: Learning Robust Visual Features without Supervision},
  author={Oquab, Maxime and Darcet, Timoth{\'e}e and Moutakanni, Th{\'e}o and Vo, Huy V and Szafraniec, Marc and Khalidov, Vasil and Fernandez, Pierre and HAZIZA, Daniel and Massa, Francisco and El-Nouby, Alaaeldin and others},
  journal={TMLR},
  year={2024}
}

@inproceedings{iclr26_RAE,
  title={Diffusion transformers with representation autoencoders},
  author={Zheng, Boyang and Ma, Nanye and Tong, Shengbang and Xie, Saining},
  booktitle=ICLR,
  pages={35791--35820},
  year={2026}
}

@article{arXiv25_VJEPA2,
  title={V-jepa 2: Self-supervised video models enable understanding, prediction and planning},
  author={Assran, Mido and Bardes, Adrien and Fan, David and Garrido, Quentin and Howes, Russell and Muckley, Matthew and Rizvi, Ammar and Roberts, Claire and Sinha, Koustuv and Zholus, Artem and others},
  journal={arXiv preprint arXiv:2506.09985},
  year={2025}
}

@inproceedings{aaai18_FiLM,
  title={Film: Visual reasoning with a general conditioning layer},
  author={Perez, Ethan and Strub, Florian and De Vries, Harm and Dumoulin, Vincent and Courville, Aaron},
  booktitle=AAAI,
  year={2018}
}

@inproceedings{icml23_BLIP2,
  title={Blip-2: Bootstrapping language-image pre-training with frozen image encoders and large language models},
  author={Li, Junnan and Li, Dongxu and Savarese, Silvio and Hoi, Steven},
  booktitle=ICML,
  pages={19730--19742},
  year={2023},
  organization={PMLR}
}

@inproceedings{iclr17_AdamW,
  title={Decoupled Weight Decay Regularization},
  author={Loshchilov, Ilya and Hutter, Frank},
  booktitle=ICLR,
  year={2017}
}

@inproceedings{iccv17_SSv2,
  title={The “something something” video database for learning and evaluating visual common sense},
  author={Goyal, Raghav and Kahou, Samira Ebrahimi and Michalski, Vincent and Materzynska, Joanna and Westphal, Susanne and Kim, Heuna and Haenel, Valentin and Fruend, Ingo and Yianilos, Peter and Mueller-Freitag, Moritz and others},
  booktitle=ICCV,
  pages={5843--5851},
  year={2017},
  organization={IEEE}
}

@inproceedings{corl22_RECON,
  title={Rapid Exploration for Open-World Navigation with Latent Goal Models},
  author={Shah, Dhruv and Eysenbach, Benjamin and Rhinehart, Nicholas and Levine, Sergey},
  booktitle=CoRL,
  pages={674--684},
  year={2022},
  organization={PMLR}
}

@article{arxiv25_LoopNav,
  title={LoopNav: Benchmarking Spatial Consistency in World Models}, 
  author={Kewei Lian and Shaofei Cai and Yitao Liang and Anji Liu},
  journal={arXiv preprint arXiv:2505.22976},
  year={2025},
}

@inproceedings{iclr23_VP2,
  title={A Control-Centric Benchmark for Video Prediction},
  author={Tian, Stephen and Finn, Chelsea and Wu, Jiajun},
  booktitle=ICLR,
  year={2023}
}

@inproceedings{corl22_BlockPushing,
  title={Implicit behavioral cloning},
  author={Florence, Pete and Lynch, Corey and Zeng, Andy and Ramirez, Oscar A and Wahid, Ayzaan and Downs, Laura and Wong, Adrian and Lee, Johnny and Mordatch, Igor and Tompson, Jonathan},
  booktitle=CoRL,
  pages={158--168},
  year={2022},
  organization={PMLR}
}

@article{ijrr25_PushT,
  title={Diffusion policy: Visuomotor policy learning via action diffusion},
  author={Chi, Cheng and Xu, Zhenjia and Feng, Siyuan and Cousineau, Eric and Du, Yilun and Burchfiel, Benjamin and Tedrake, Russ and Song, Shuran},
  journal={IJRR},
  volume={44},
  number={10-11},
  pages={1684--1704},
  year={2025},
  publisher={Sage Publications Sage UK: London, England}
}

@inproceedings{icra16_MPPI,
  title={Aggressive driving with model predictive path integral control},
  author={Williams, Grady and Drews, Paul and Goldfain, Brian and Rehg, James M and Theodorou, Evangelos A},
  booktitle={ICRA},
  pages={1433--1440},
  year={2016},
  organization={IEEE}
}

@article{arXiv20_STT,
  title={Spatial-Temporal Transformer Networks for Traffic Flow Forecasting}, 
  author={Mingxing Xu and Wenrui Dai and Chunmiao Liu and Xing Gao and Weiyao Lin and Guo-Jun Qi and Hongkai Xiong},
  journal={arXiv preprint arXiv:2001.02908},
  year={2021},
}

@inproceedings{icml19_SetTransformer,
  title={Set transformer: A framework for attention-based permutation-invariant neural networks},
  author={Lee, Juho and Lee, Yoonho and Kim, Jungtaek and Kosiorek, Adam and Choi, Seungjin and Teh, Yee Whye},
  booktitle=ICML,
  pages={3744--3753},
  year={2019},
  organization={PMLR}
}

@inproceedings{eccv20_DETR,
  title={End-to-end object detection with transformers},
  author={Carion, Nicolas and Massa, Francisco and Synnaeve, Gabriel and Usunier, Nicolas and Kirillov, Alexander and Zagoruyko, Sergey},
  booktitle=ECCV,
  pages={213--229},
  year={2020},
  organization={Springer}
}

@inproceedings{neurips22_Flamingo,
  title={Flamingo: a visual language model for few-shot learning},
  author={Alayrac, Jean-Baptiste and Donahue, Jeff and Luc, Pauline and Miech, Antoine and Barr, Iain and Hasson, Yana and Lenc, Karel and Mensch, Arthur and Millican, Katherine and Reynolds, Malcolm and others},
  booktitle=NeurIPS,
  volume={35},
  pages={23716--23736},
  year={2022}
}

@inproceedings{icml23_RankMe,
  title = {RankMe: Assessing the Downstream Performance of Pretrained Self Supervised Representations by Their Rank},
  author = {Quentin Garrido and Randall Balestriero and Laurent Najman and Yann LeCun},
  booktitle = ICML,
  pages = {10929--10974},
  publisher = {PMLR},
  year = {2023},
}

@article{arxiv18_UMAP,
  title={Umap: Uniform manifold approximation and projection for dimension reduction},
  author={McInnes, Leland and Healy, John and Melville, James},
  journal={arXiv preprint arXiv:1802.03426},
  year={2018}
}

@inproceedings{icml18_Procgen,
  title={Leveraging procedural generation to benchmark reinforcement learning},
  author={Cobbe, Karl and Hesse, Chris and Hilton, Jacob and Schulman, John},
  booktitle=ICML,
  pages={2048--2056},
  year={2020},
  organization={PMLR}
}

@inproceedings{cvpr16_CAM,
  title = {Learning Deep Features for Discriminative Localization},
  author = {Bolei Zhou and Aditya Khosla and Agata Lapedriza and Aude Oliva and Antonio Torralba},
  booktitle = CVPR,
  pages = {2921--2929},
  year = {2016},
}

@article{tip04_SSIM,
  title={Image quality assessment: from error visibility to structural similarity},
  author={Wang, Zhou and Bovik, Alan C and Sheikh, Hamid R and Simoncelli, Eero P},
  journal=TIP,
  volume={13},
  number={4},
  pages={600--612},
  year={2004},
  publisher={IEEE}
}

@inproceedings{cvpr18_LPIPS,
  title={The unreasonable effectiveness of deep features as a perceptual metric},
  author={Zhang, Richard and Isola, Phillip and Efros, Alexei A and Shechtman, Eli and Wang, Oliver},
  booktitle=CVPR,
  pages={586--595},
  year={2018},
  organization={IEEE}
}
\bibliographystyle{iclr2027_conference}

\newpage
\appendix
\makeatletter
\setlength{\@fptop}{0pt}
\makeatother
\section*{Appendix}

\begin{center}
\large\textbf{Appendix Contents}
\end{center}
\vspace{0.8em}
\begingroup
\normalsize
\appxtosec{sec:appx:details}{A \quad Implementation Details}
\appxtosubsec{sec:appx:config}{\quad A.1 \quad Model Settings}
\appxtosubsec{sec:appx:hyper}{\quad A.2 \quad Training Hyperparameters}
\appxtosubsec{sec:appx:params}{\quad A.3 \quad Parameters Comparison}
\appxtosubsec{sec:appx:models}{\quad A.4 \quad Model Descriptions}
\appxtosubsec{sec:appx:datasets}{\quad A.5 \quad Datasets}

\appxtosec{sec:appx:results}{B \quad More Results}
\appxtosubsec{sec:appx:lp}{\quad B.1 \quad Linear Probing}
\appxtosubsec{sec:appx:diversity}{\quad B.2 \quad Action Diversity}
\appxtosubsec{sec:appx:umaps}{\quad B.3 \quad UMAP Visualization}
\appxtosubsec{sec:appx:a2p}{\quad B.4 \quad Action-to-Patch Attention}
\appxtosubsec{sec:appx:rollout}{\quad B.5 \quad Reconstruction MSE}
\appxtosubsec{sec:appx:gene}{\quad B.6 \quad Video Generation}
\appxtosubsec{sec:appx:action_transfer}{\quad B.7 \quad Action Transfer}
\appxtosubsec{sec:appx:util}{\quad B.8 \quad Utilization Analysis}
\appxtosubsec{sec:appx:vp}{\quad B.9 \quad Visual Planning}

\appxtosec{sec:appx:efficiency}{C \quad Efficiency Evaluation}

\appxtosec{sec:appx:analyses}{D \quad More Analyses}
\appxtosubsec{sec:appx:lam_pretraining}{\quad D.1 \quad LAM Pretraining}
\appxtosubsec{sec:appx:vp2_finetuning}{\quad D.2 \quad VP$^2$ Finetuning}
\appxtosubsec{sec:appx:vp2_ablation}{\quad D.3 \quad VP$^2$ Ablation}
\appxtosubsec{sec:appx:vp2_video}{\quad D.3 \quad VP$^2$ Video Prediction}

\endgroup
\clearpage

\appendix
\section{Implementation Details}
\label{sec:appx:details}
\subsection{Model Settings}
\label{sec:appx:config}
Our ACT-LAM has only about 55M trainable parameters. Following DiLA \citep{icml26_DiLA}, we employ the DINOv2 and the pretrained ViT-XL from RAE as the visual encoder and decoder, respectively. The default settings are detailed in Tab.~\ref{tab:appx:model_config}. 

\subsection{Training Hyperparameters}
\label{sec:appx:hyper}
All experiments are implemented in the PyTorch framework and conducted on \textbf{four NVIDIA RTX A6000 GPUs}, each equipped with 48\,GB of memory. we train all models in full precision (FP32) and using AdamW with a learning rate of $1 \times 10^{-4}$, $(\beta_1,\beta_2)=(0.9,0.999)$, and a weight decay of $1 \times 10^{-5}$. We apply gradient clipping with a maximum norm of 1.0 and use a constant learning rate without warm-up or decay. Input videos are resized to a resolution of $256 \times 256$ and sampled with a sequence length of 16 frames. The model is trained under a teacher-forcing regime for 50k iterations. The loss coefficients are set to $\lambda_{\mathrm{r}}=2.0$, $\lambda_{\mathrm{g}}=0.05$, $\lambda_{\mathrm{a}}=0.05$, 
$\lambda_{\mathrm{v}}=0.01$, and
$\lambda_{\mathrm{s}}=0.001$. 

\subsection{Parameters Comparison}
\label{sec:appx:params}
We report the parameter sizes of several latent action models in Tab.~\ref{tab:appx:params}. ACT-LAM uses only about half as many trainable parameters as DiLA \citep{icml26_DiLA}, reducing the memory demand during training and enabling deployment on GPUs with smaller memory capacity.

\vspace{-1em}
\begin{table*}[tbhp]
\centering
\caption{\textbf{Model parameters.}}
\vspace{0.5pt}
\begin{center}
\begin{sc}
\renewcommand{\arraystretch}{1.3}
\resizebox{0.8\textwidth}{!}{
\begin{tabular}{l c c c c c}
    \toprule
    \bf Parameters & ACT-LAM & DiLA & AdaWorld(LAM) & AdaWorld & Genie\\
    \midrule
    \bf Trainable & 55M & 123M & 500M & 1.5B & 11B \\
    \bf Frozen & 500M & 500M & - & - & - \\
    \bottomrule
\end{tabular}}
\end{sc}
\end{center}
\label{tab:appx:params}
\vspace{-1em}
\end{table*}

\subsection{Model Descriptions}
\label{sec:appx:models}
We provide detailed descriptions of the LAMs used in Fig.~\ref{fig:motivation}. \emph{DiLA} \citep{icml26_DiLA} serves as a representative LAM. It employs a feature processing pathway to construct compact structure representations before latent action inference, together with a DeepConv IDM and an FDM for latent dynamics modeling. \emph{CoMo} \citep{cvpr26_CoMo} instead infers continuous latent actions from rich visual representations using Motion Q-Former, where learnable motion queries interact with visual tokens through global self-attention.

To better analyze the relation between reconstruction and latent dynamics, we further construct two controlled models: (i) \emph{PlainLAM} removes the content pathway from \emph{DiLA} while retaining its main latent dynamics pathway. This model eliminates the additional content modeling and therefore provides a cleaner baseline \emph{PlainLAM w/o STT} for examining whether improved reconstruction corresponds to better latent dynamics. (ii) \emph{MFormer} further replaces the DeepConv IDM with the Motion Q-Former used in CoMo, while removing the ST-Transformer block. This allows us to examine the effect of query learning independently of other designs.

\subsection{Datasets}
\label{sec:appx:datasets}
\noindent{\textbf{SSv2.}} 
Something-Something v2 (SSv2) \citep{iccv17_SSv2} is a large-scale video dataset consisting of short clips of humans performing with everyday objects. It contains $220,847$ videos spanning $174$ action categories, with $168,913$ videos for training, $24,777$ for validation, and $27,157$ for testing. In contrast to conventional action recognition datasets that the scene appearance can provide strong class cues, SSv2 emphasizes fine-grained object interactions and temporal motion patterns. This property makes it particularly suitable for latent action learning, as the visual transitions contain rich action-related motion cues. In our experiments, we use SSv2 as a source of human interaction videos for pretraining LAMs. 

\noindent{\textbf{RT-1.}} 
RT-1 \citep{arxiv22_RT1} is a real-world robot manipulation dataset collected with mobile manipulators performing a diverse set of tasks. The original RT-1 contains over $130$K episodes covering more than $700$ tasks. Following DiLA, we use the subset released through Open X-Embodiment, which contains $87,212$ training episodes. Each episode provides temporally aligned visual observations and robot interactions, offering state transitions under realistic physical dynamics. We use RT-1 to expose the latent action models to embodied manipulation behaviors that complement the human-object interactions in SSv2.

\begin{table}[t]
\caption{\bf Default model settings.}
\vspace{1pt}
\label{tab:appx:model_config}
\begin{center}
\renewcommand{\arraystretch}{1.1}
\scalebox{1.0}{
\begin{tabular}{lc}
\toprule
\multicolumn{1}{c}{\bf COMPONENT/PARAMETER} &
\multicolumn{1}{c}{\bf VALUE}
\\ \midrule
\multicolumn{2}{l}{\bf Input Parameters} \\
Input Image dimensions & $256 \times 256 \times 3$ \\
DINOv2 embedding dimension & 768 \\
Training frame length & 16 \\
Spatial patch numbers & $16 \times 16$ \\ \hline
\multicolumn{2}{l}{\bf State Adapter (MLP)} \\
Input per-patch dimension & 768 \\
Hidden dimension & 384 \\
Output per-patch state dimension & 128 \\
Residual MLP block depth & 3 \\ \hline
\multicolumn{2}{l}{\bf IDM (Action Query IDM)} \\
Input transition dimension & 128 \\
Hidden dimension & 384 \\
Latent Action Dimension (Global) $d_z$ & 256 \\
Learned Action Query numbers & 8 \\
Attention heads & 8 \\
Head dimension & 48 \\
Local 3D transition block depth & 3 \\
Local 3D convolution kernel & $3 \times 3 \times 3$ \\
Query Mixer layers & 2 \\
Temporal Mixer layers & 2 \\
Temporal context radius & 1 \\
Learned spatial position & $16 \times 16$ \\ \hline
\multicolumn{2}{l}{\bf FDM (Action-Token FDM)} \\
Input per-patch state dimension & 128 \\
Per-patch hidden dimension & 384 \\
Output per-patch state residual dimension & 128 \\
Depth & 4 \\
Heads numbers & 8 \\
Head dimension & 48 \\
FFN multiplier & 4 \\
Learned spatial position & $16 \times 16$ \\
Learned temporal position length & 64 \\ \hline
\multicolumn{2}{l}{\bf Initial-Frame Fusion Decoder} \\
Input state dimension & 128 \\
Reference frame dimension & 768 \\
Structure hidden dimension & 768 \\
Decoder hidden dimension & 768 \\
Structure decoder depth & 2 \\
Fusion block numbers & 2 \\
Attention heads & 8 \\
\bottomrule
\end{tabular}}
\end{center}
\end{table}

\noindent{\textbf{RECON.}}
RECON \citep{corl22_RECON} is a visual navigation dataset collected using mobile ground robots in diverse outdoor environments. The dataset contains over $5,000$ trajectories recorded across $9$ environments. In addition to RGB observations, the original release includes measurements from multiple sensors. Following DiLA \citep{icml26_DiLA}, the trajectories are organized into training and test splits, and only the training set is used for LAM pretraining. Compared with RT-1, RECON contains continuous ego-motion and navigation dynamics, providing a challenging setting for modeling continuous motion manifolds.

\noindent{\textbf{LoopNav.}}
LoopNav \citep{arxiv25_LoopNav} is a large-scale navigation dataset collected in the open-world Minecraft environment. It contains approximately $250$ hours of navigation trajectories, corresponding to about $20$ million frames, together with temporally aligned actions. The control interface restricts each time step to a single atomic action, providing a relatively clean correspondence between visual transitions and agent motion. Due to the scale of the full dataset, we utilize a subset for our experiments: Villages and Biomes $\&$ Locations. 

\noindent{\textbf{VP$^2$.}}
VP$^2$ \citep{iclr23_VP2} is a control-centric benchmark for evaluating video prediction models through downstream robotic manipulation. The full benchmark consists of $2$ simulated environments, RoboSuite and RoboDesk, covering $11$ manipulation task categories and $310$ task instances in total. Specifically, RoboSuite contains $4$ tabletop pushing categories, while RoboDesk contains $7$ manipulation categories. Following the evaluation protocol adopted by AdaWorld \citep{icml25_AdaWorld} and DiLA \citep{icml26_DiLA}, the $4$ RoboSuite categories are aggregated as \emph{RoboSuite Push}, and we evaluate the $5$ RoboDesk tasks reported by them. In our experiments, we use VP$^{2}$ to assess whether the latent actions learned by ACT-LAM translate into improved downstream visual planning performance.

\noindent{\textbf{Block Pushing.}}
Block Pushing \citep{corl22_BlockPushing} is a simulated manipulation benchmark in which a robot interacts with two colored blocks and two target regions. The dataset contains $1,000$ trajectories generated by a scripted controller, with randomized initial block positions and multiple valid manipulation strategies. Each trajectory provides state transitions paired with continuous control actions. Block Pushing serves as a useful testbed for examining whether latent action representations preserve the continuous control signals beyond the pretraining domains.

\noindent{\textbf{Push-T.}}
Push-T \citep{ijrr25_PushT} is a manipulation benchmark that requires to push a T-shaped object toward a fixed target pose from randomized initial configurations. The simulated dataset contains $200$ expert demonstrations and provides synchronized visual observations and two-dimensional continuous actions. Successful manipulation requires coordinated contact and precise object motion, while multiple action trajectories can lead to similar goal states. These properties provide a complementary setting for evaluating whether learned latent actions encode fine-grained continuous manipulation behaviors.

\begin{table*}[h]
\centering
\caption{\textbf{Linear probing MSE.} Results represent the mean and standard deviation of 4 independent runs per task. Baseline adopts the IDM and FDM used in DiLA. The best results are in ``\textbf{bold}".}
\vspace{0pt}
\label{tab:appx:lp}
\begin{center}
\renewcommand{\arraystretch}{1.05}
\resizebox{0.6\textwidth}{!}{
\begin{tabular}{c c c c}
    \toprule
    \textbf{Method} & Block Pushing 
    & Push-T\\
    \midrule
    {PlainLAM} 
    & \tabnum{0.103\pm0.010} 
    & \tabnum{0.037\pm0.001} \\
    
    {PlainLAM w/o STT} 
    & \tabnum{0.104\pm0.011} 
    & \tabnum{0.037\pm0.001} \\
    \midrule

    {Baseline} 
    & \tabnum{0.061\pm0.006} 
    & \tabnum{0.018\pm0.000} \\

    {Baseline w/ AQ-IDM} 
    & \tabnum{0.035\pm0.004} 
    & \tabnum{0.011\pm0.000} \\

    {Baseline w/ AT-FDM} 
    & \tabnum{0.050\pm0.010} 
    & \tabnum{0.014\pm0.001} \\
    
    {ACT-LAM w/o $\mathcal{L}_{var}$} 
    & \bm{\tabnum{0.033\pm0.005}} 
    & \bm{\tabnum{0.010\pm0.000}} \\
    \bottomrule
\vspace{-2.5em}
\end{tabular}}
\end{center}
\end{table*}

\section{More Results}
\label{sec:appx:results}
\subsection{Linear Probing}
\label{sec:appx:lp}
To evaluate the quality of the learned latent actions, we perform linear probing on Block Pushing \citep{corl22_BlockPushing} and Push-T \citep{ijrr25_PushT}. For each benchmark, we use $1,000$ transition pairs with $16$ frames: $800$ pairs are used for training and $200$ pairs for testing. We optimize the MSE using SGD with learning rate $0.01$, batch size $32$, and $5,000$ epochs. The probe is repeated with four independent seeds. More results are shown in Tab.~\ref{tab:appx:lp}. 

\begin{figure*}[t]
  \centering
   \includegraphics[width=1.0\linewidth]{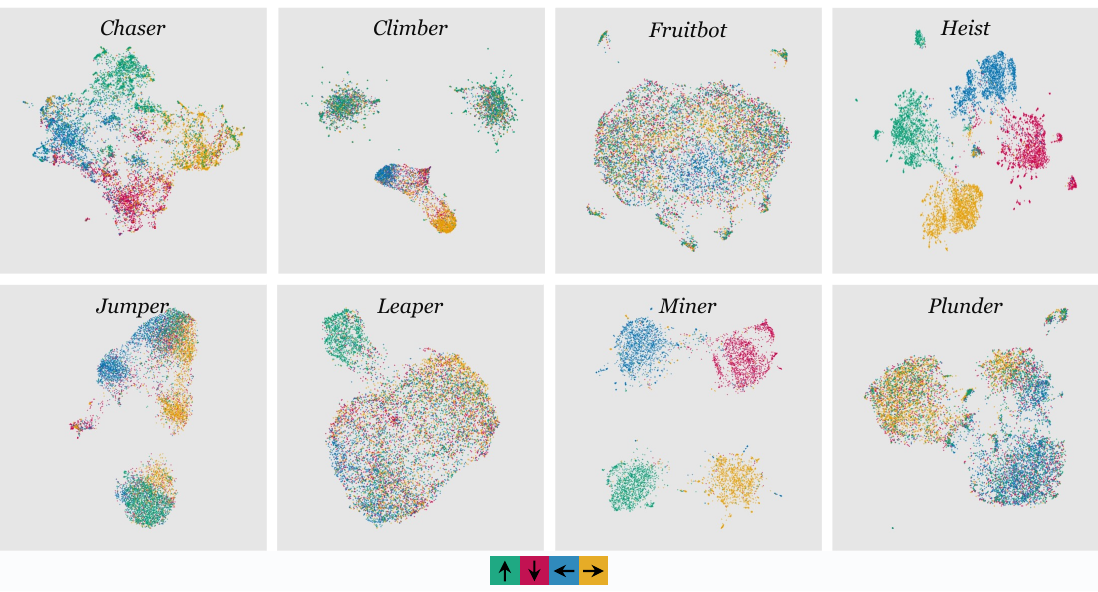}
    \caption{\textbf{UMAP projection of the learned latent action space.} We visualize the latent action spaces of additional $8$ Procgen environments.}
   \label{fig:appx:umaps}
   \vspace{-1em}
\end{figure*}

\begin{figure}[t]
  \centering
   \includegraphics[width=1.0\linewidth]{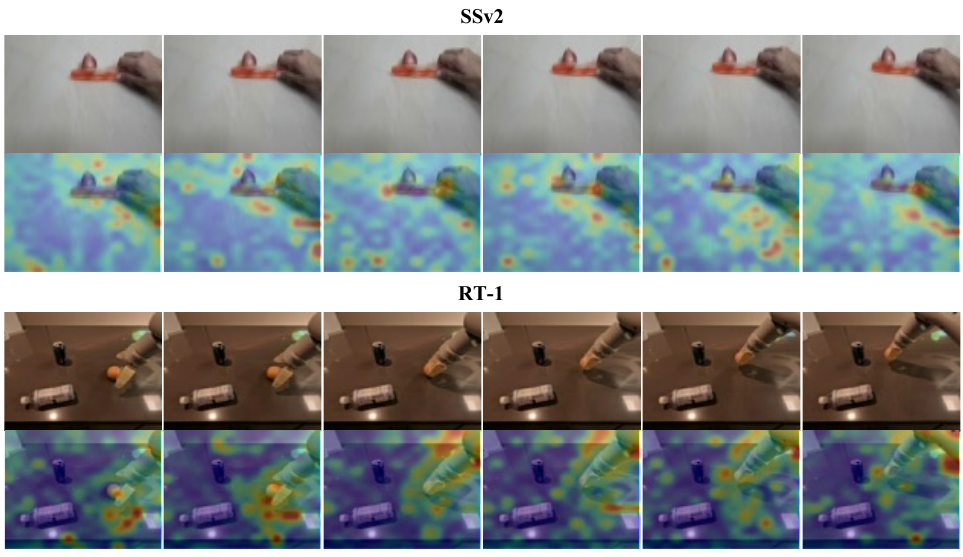}
    \caption{\textbf{Additional action-to-patch attention visualizations on SSv2 and RT-1.} The top rows show the input frames, while the bottom rows show the corresponding attention heatmaps produced by AQ-IDM. The attention maps consistently highlight action-related regions.}
   \label{fig:appx:a2ps}
   \vspace{-1em}
\end{figure}

\subsection{Action Diversity}
\label{sec:appx:diversity}
Following \cite{icml23_RankMe}, we report the logarithm of singular values normalized by the largest singular value. Specifically, for each LAM, we sample $1,000$ RGB transition sequences, each consisting of $16$ frames. For each LAM, the IDM produces $15$ latent actions from each context. We then select the target transition and obtain the latent action matrix $\mathcal{Z} \in \mathbb{R}^{1000\times d_z}$. After that, we center $\mathcal{Z}$ across samples and compute its singular values $\sigma_1\geq\cdots\geq\sigma_r$. We report the normalized log singular-value spectrum $\log(\sigma_i/\sigma_1)$ in Fig.~\ref{fig:singular}. Notably, a slower spectral decay indicates that the latent actions span a broader set of independent transition directions, reflecting higher diversity.

\subsection{UMAP Visualization}
\label{sec:appx:umaps}
We visualize the latent action spaces on selected test splits of the Procgen benchmark \citep{icml18_Procgen}. For each environment, we use a set of $10,000$ transitions, with $2,500$ samples from each of the four cardinal action classes, \emph{i.e.}, \texttt{LEFT}, \texttt{RIGHT}, \texttt{UP}, and \texttt{DOWN}. Ground-truth actions are used only for color coding in the visualization. We include only environments with sufficient samples from all four action classes. Additional visualizations are provided in Fig.~\ref{fig:appx:umaps}.

\begin{figure*}[t]
  \centering
   \includegraphics[width=1.0\linewidth]{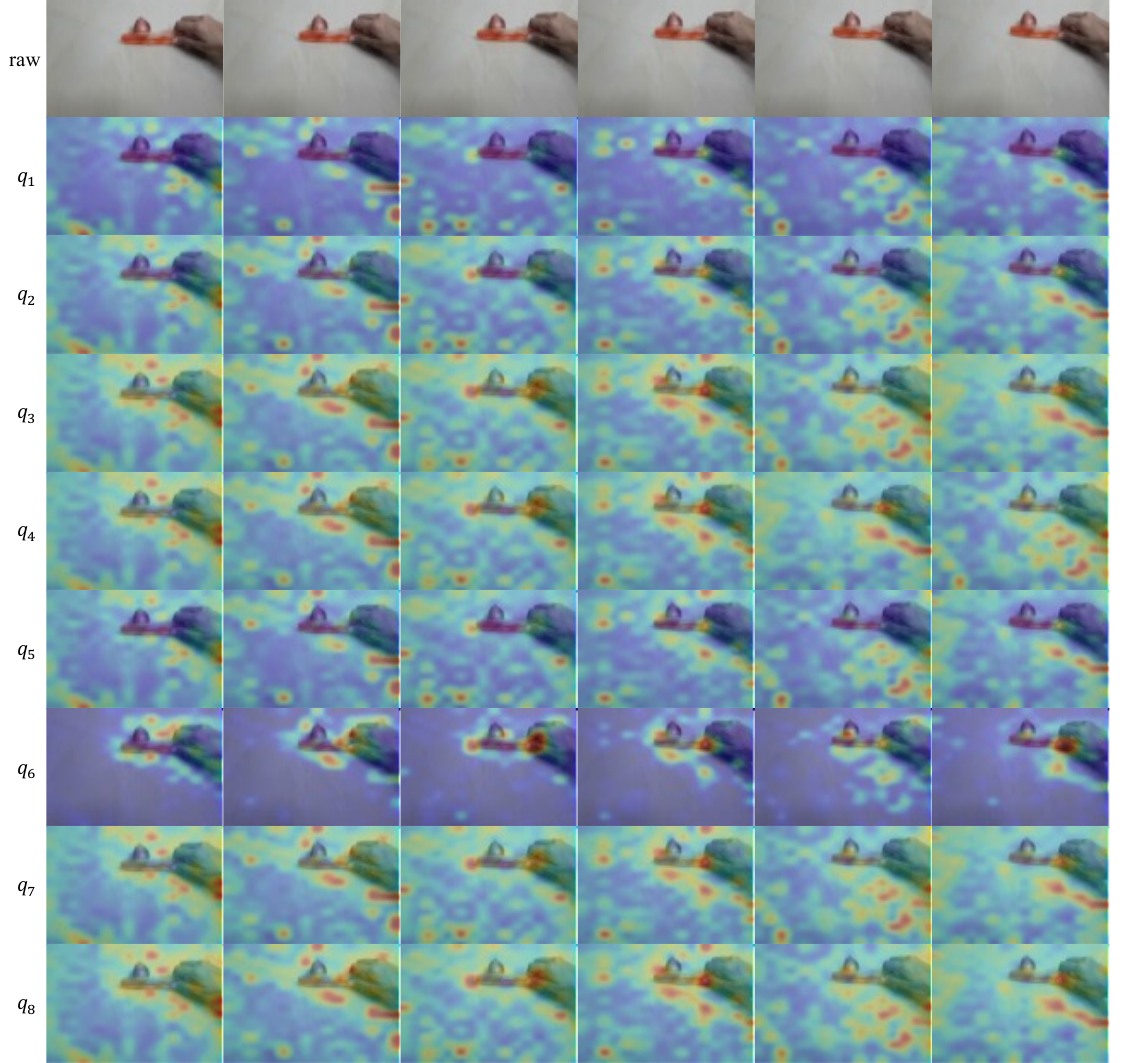}
    \caption{\textbf{Per-query action-to-patch attention maps.} The top row shows the input RGB frames, while the remaining rows visualize the attention heatmaps produced by individual action queries $q_1$ to $q_8$. Most queries distribute their attention broadly across action-related regions, while $q_6$ shows a markedly more localized attention pattern.}
   \label{fig:appx:per_q}
   \vspace{-1em}
\end{figure*}

\subsection{Action-to-Patch Attention}
\label{sec:appx:a2p}
We provide additional visualizations of the action-to-patch attention \citep{cvpr16_CAM} in AQ-IDM. Specifically, we extract the query-to-patch attention weights and reshape them according to the spatial patch layout. Then, we upsample the resulting maps to the input image resolution for visualization. Fig.~\ref{fig:appx:a2ps} presents additional examples from SSv2 and RT-1. To further observe the behavior of individual action queries, Fig.~\ref{fig:appx:per_q} visualizes the attention maps of $q_1$ to $q_8$ separately on an SSv2 sequence. Most queries distribute their attention broadly across action-related regions, while $q_6$ shows a markedly more localized attention pattern. These results provide a more detailed view of how individual action queries respond to spatial transition cues.

\vspace{-1em}
\begin{table}[h]
\centering
\caption{\textbf{More rollout results.}}
\label{tab:appx:rollout_mse}
\vspace{0.3em}
\setlength{\tabcolsep}{5pt}
\renewcommand{\arraystretch}{1.3}
\resizebox{0.8\textwidth}{!}{
\begin{tabular}{cccccc}
\toprule
\multirow{2}{*}{\textbf{Dataset}}
& \multirow{2}{*}{\textbf{Model}}
& \multicolumn{4}{c}{\textbf{Recon. MSE} \ $\bm{\downarrow}$} \\
\cmidrule(lr){3-6}
& & @1 & @4 & @8 & @15 \\
\midrule
\multirow{2}{*}{SSv2}
& Baseline
& \tabnum{0.174\pm0.087}
& \bm{\tabnum{0.285\pm0.114}} 
& \bm{\tabnum{0.451\pm0.119}}
& \bm{\tabnum{0.656\pm0.109}} \\
& Baseline w/ AT-FDM
& \bm{\tabnum{0.172\pm0.085}}
& \tabnum{0.286\pm0.109}
& \tabnum{0.516\pm0.097}
& \tabnum{0.788\pm0.076} \\
\midrule
\multirow{2}{*}{RT-1}
& Baseline
& \tabnum{0.120\pm0.053}
& \bm{\tabnum{0.218\pm0.080}}
& \bm{\tabnum{0.379\pm0.082}}
& \bm{\tabnum{0.579\pm0.069}} \\
& Baseline w/ AT-FDM
& \bm{\tabnum{0.118\pm0.052}}
& \tabnum{0.227\pm0.073}
& \tabnum{0.490\pm0.068}
& \tabnum{0.749\pm0.057} \\
\bottomrule
\end{tabular}}
\end{table}

\vspace{-1.5em}
\begin{table}[h]
\centering
\caption{\textbf{Teacher-forcing results.}}
\label{tab:appx:tf}
\vspace{0.3em}
\setlength{\tabcolsep}{8pt}
\renewcommand{\arraystretch}{1.3}
\resizebox{0.48\textwidth}{!}{
\begin{tabular}{ccc}
\toprule
\multirow{2}{*}{\textbf{Model}}
& \multicolumn{2}{c}{\textbf{Recon. MSE} $\bm{\downarrow}$} \\
\cmidrule(lr){2-3}
& \textbf{SSv2} & \textbf{RT-1} \\
\midrule
DiLA
& \tabnum{0.386\pm0.113}
& \tabnum{0.236\pm0.061} \\
Baseline
& \tabnum{0.168\pm0.066}
& \tabnum{0.121\pm0.035} \\
Baseline w/ AT-FDM
& \bm{\tabnum{0.166\pm0.065}}
& \bm{\tabnum{0.118\pm0.034}} \\
ACT-LAM
& \tabnum{0.169\pm0.064}
& \tabnum{0.120\pm0.034} \\
\bottomrule
\end{tabular}}
\end{table}
\vspace{-1em}

\begin{table}[t]
\centering
\caption{\textbf{Rollout SSIM on SSv2 and RT-1.}}
\label{tab:appx:ssim}
\vspace{0.3em}
\setlength{\tabcolsep}{7pt}
\renewcommand{\arraystretch}{1.28}
\resizebox{0.8\linewidth}{!}{%
\begin{tabular}{llcccc}
\toprule
\multirow{2}{*}{\textbf{Dataset}}
& \multirow{2}{*}{\textbf{Model}}
& \multicolumn{4}{c}{\textbf{SSIM} $\uparrow$} \\
\cmidrule(lr){3-6}
& & @1 & @4 & @8 & @15 \\
\midrule
\multirow{4}{*}{SSv2}
& \color{gray!80}\textit{Recon.}
& \graynum{0.718\pm0.142}
& \graynum{0.717\pm0.141}
& \graynum{0.717\pm0.140}
& \graynum{0.717\pm0.141} \\
& AdaWorld (FDM)
& \bm{\tabnum{0.692\pm0.154}}
& \tabnum{0.611\pm0.157}
& \tabnum{0.580\pm0.158}
& \tabnum{0.557\pm0.160} \\
& DiLA
& \tabnum{0.613\pm0.149}
& \tabnum{0.599\pm0.148}
& \bm{\tabnum{0.588\pm0.148}}
& \bm{\tabnum{0.578\pm0.150}} \\
& ACT-LAM
& \tabnum{0.678\pm0.145}
& \bm{\tabnum{0.633\pm0.145}}
& \tabnum{0.572\pm0.147}
& \tabnum{0.508\pm0.148} \\
\midrule
\multirow{4}{*}{RT-1}
& \color{gray!80}\textit{Recon.}
& \graynum{0.623\pm0.075}
& \graynum{0.622\pm0.073}
& \graynum{0.621\pm0.073}
& \graynum{0.621\pm0.072} \\
& AdaWorld (FDM)
& \bm{\tabnum{0.679\pm0.091}}
& \tabnum{0.500\pm0.095}
& \tabnum{0.424\pm0.083}
& \tabnum{0.377\pm0.074} \\
& DiLA
& \tabnum{0.565\pm0.078}
& \bm{\tabnum{0.545\pm0.080}}
& \bm{\tabnum{0.526\pm0.081}}
& \bm{\tabnum{0.506\pm0.082}} \\
& ACT-LAM
& \tabnum{0.589\pm0.075}
& \tabnum{0.536\pm0.072}
& \tabnum{0.449\pm0.067}
& \tabnum{0.375\pm0.060} \\
\bottomrule
\end{tabular}}
\vspace{-0.8em}
\end{table}

\begin{table}[h]
\centering
\caption{\textbf{Rollout LPIPS on SSv2 and RT-1.}}
\label{tab:appx:lpips}
\vspace{0.3em}
\setlength{\tabcolsep}{7pt}
\renewcommand{\arraystretch}{1.28}
\resizebox{0.8\linewidth}{!}{%
\begin{tabular}{llcccc}
\toprule
\multirow{2}{*}{\textbf{Dataset}}
& \multirow{2}{*}{\textbf{Model}}
& \multicolumn{4}{c}{\textbf{LPIPS} $\downarrow$} \\
\cmidrule(lr){3-6}
& & @1 & @4 & @8 & @15 \\
\midrule
\multirow{4}{*}{SSv2}
& \color{gray!80}\textit{Recon.}
& \graynum{0.125\pm0.040}
& \graynum{0.125\pm0.038}
& \graynum{0.125\pm0.038}
& \graynum{0.125\pm0.038} \\
& AdaWorld (FDM)
& \tabnum{0.436\pm0.184}
& \tabnum{0.612\pm0.158}
& \tabnum{0.678\pm0.142}
& \tabnum{0.723\pm0.130} \\
& DiLA
& \tabnum{0.387\pm0.149}
& \tabnum{0.489\pm0.143}
& \tabnum{0.568\pm0.136}
& \tabnum{0.637\pm0.130} \\
& ACT-LAM
& \bm{\tabnum{0.191\pm0.084}}
& \bm{\tabnum{0.231\pm0.095}}
& \bm{\tabnum{0.316\pm0.096}}
& \bm{\tabnum{0.430\pm0.087}} \\
\midrule
\multirow{4}{*}{RT-1}
& \color{gray!80}\textit{Recon.}
& \graynum{0.146\pm0.027}
& \graynum{0.146\pm0.024}
& \graynum{0.146\pm0.024}
& \graynum{0.147\pm0.023} \\
& AdaWorld (FDM)
& \tabnum{0.315\pm0.114}
& \tabnum{0.560\pm0.122}
& \tabnum{0.656\pm0.102}
& \tabnum{0.713\pm0.090} \\
& DiLA
& \tabnum{0.220\pm0.056}
& \tabnum{0.259\pm0.072}
& \bm{\tabnum{0.305\pm0.098}}
& \bm{\tabnum{0.364\pm0.125}} \\
& ACT-LAM
& \bm{\tabnum{0.177\pm0.036}}
& \bm{\tabnum{0.222\pm0.050}}
& \tabnum{0.325\pm0.053}
& \tabnum{0.438\pm0.049} \\
\bottomrule
\end{tabular}}
\vspace{-0.8em}
\end{table}

\subsection{Reconstruction MSE}
\label{sec:appx:rollout}
We evaluate multi-step forward dynamics on $2,000$ videos from the SSv2 test split and $2,000$ videos from the RT-1 validation split, using the same evaluation set across all methods. Given an initial state and a sequence of latent actions, each model auto-regressively predicts subsequent states by harnessing its predicted state into the next step. The prediction errors are measured by MSE between the predicted and ground-truth DINO features. For each video, we average the per-step MSE over the first $k$ predicted steps, with $k \in \{1,4,8,15\}$. We then report the mean and standard deviation of the videos in each dataset. Notably, DiLA \citep{icml26_DiLA} is evaluated using its official implementation and checkpoint, which includes rollout finetuning. ACT-LAM is evaluated directly after teacher-forcing pretraining, without additional finetuning. More results are in Tab.~\ref{tab:appx:rollout_mse} and \ref{tab:appx:tf}.

\subsection{Video Generation}
\label{sec:appx:gene}
We evaluate video generation quality on the same samples used for the MSE evaluation in Sec.~\ref{sec:appx:rollout}. Each model generates 15 future frames autoregressively, and SSIM \citep{tip04_SSIM} and LPIPS \citep{cvpr18_LPIPS} are computed against the raw RGB frames. @$k$ denotes the averaged metric over the first $k$ predicted frames. DiLA \citep{icml26_DiLA} and AdaWorld \citep{icml25_AdaWorld} use their official released code and checkpoints. {\color{gray!80}\textit{Recon.}} passes each ground-truth future frame directly through the frozen DINOv2 encoder \cite{tmlr24_DINOv2} and RAE decoder \citep{iclr26_RAE} without invoking latent
actions or FDM rollout, and thus serves as a reconstruction reference.

As shown in Tab.~\ref{tab:appx:ssim} and~\ref{tab:appx:lpips}, ACT-LAM achieves the best LPIPS across all SSv2 horizons and at short horizons on RT-1, while maintaining comparable SSIM performance. DiLA performs better in long-horizon SSIM, suggesting stronger structural alignment after rollout finetuning. SSIM emphasizes local structural consistency whereas LPIPS measures perceptual similarity. The superior LPIPS performance of ACT-LAM indicates that its action-conditioned dynamics better preserve perceptually plausible visual content. 

\begin{figure*}[t]
  \centering
   \includegraphics[width=1.0\linewidth]{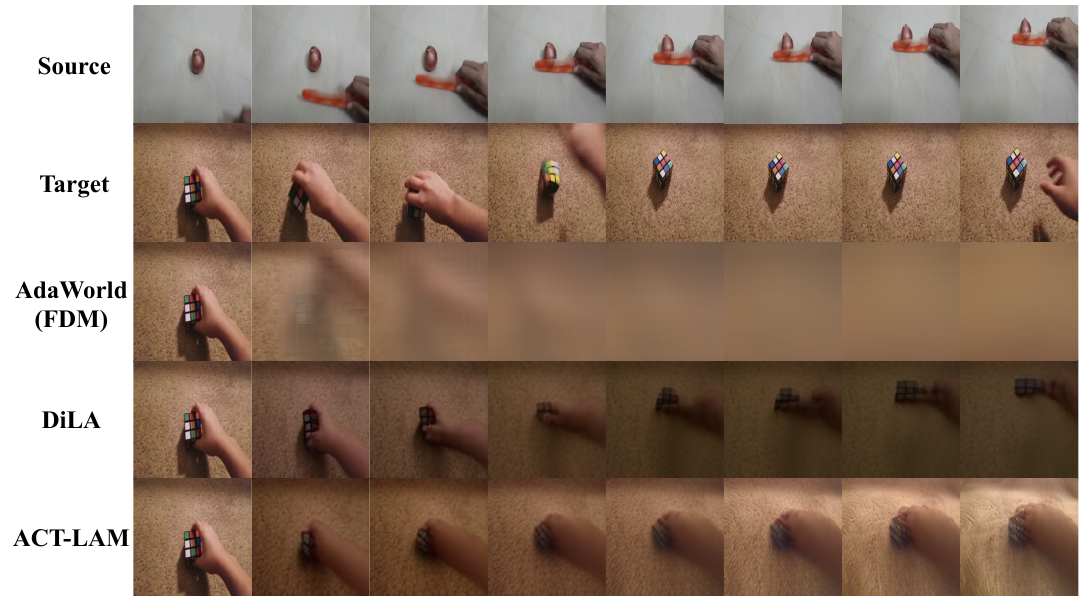}
    \caption{\textbf{Qualitative comparison of action transferability on SSv2.} We extract a 16-step latent action sequence from the source video and apply it autoregressively to the target visual context. The top two rows show the source and target reference sequences, while the remaining rows show the transferred rollouts generated by AdaWorld, DiLA, and ACT-LAM.}
   \label{fig:appx:transfer}
   \vspace{-1em}
\end{figure*}

\subsection{Action Transfer}
\label{sec:appx:action_transfer}
We evaluate action transferability by extracting a 16-step action sequence from a source video and applying it to a different visual context through autoregressive generation. Fig.~\ref{fig:appx:transfer} presents the qualitative comparisons among AdaWorld, DiLA, and ACT-LAM. DiLA generally achieves the highest visual fidelity, while ACT-LAM still captures motion patterns across different contexts. We emphasize that this experiment is intended to assess the transferability of the learned latent actions rather than reconstruction fidelity, which is not the primary objective of ACT-LAM.

\subsection{Utilization Analysis}
\label{sec:appx:util}
\noindent{\textbf{Definition.}}
To quantify latent action utilization, we compare two rollouts: (i) driven by the original latent action sequence, and (ii) using a constant action at every step. Specifically, the FDM predicts a state residual with Eq.~(\ref{eq:fdm}). For a horizon $h$, we compute:
\begin{equation}
    E_{\mathrm{action}}(h)= \frac{1}{h}\sum_{i=1}^{h}\operatorname{MSE}(\hat{s}_{i},s_i).
\end{equation}

And we obtain $E_{\mathrm{constant}}(h)$ by replacing every $z_t$
with the mean action $c$ of the training set. Based on $E_{\mathrm{action}}(h)$ and $E_{\mathrm{constant}}(h)$, we define action utilization as:

\begin{equation}
    U(h)=\frac{E_{\mathrm{constant}}(h)-E_{\mathrm{action}}(h)}{E_{\mathrm{constant}}(h)}\times 100\%.
\end{equation}

Thus, $U(h)$ measures the state MSE reduction achieved by the original latent actions relative to the constant action baseline. We deem that higher values indicate more effective action utilization.

\noindent{\textbf{Experimental Protocol.}}
We freeze the pretrained visual encoder, state adapter, and AQ-IDM, and train only different FDMs. For each dataset, we use the same fixed $8,000$/$1,000$/$1,000$ split for training, validation, and testing, respectively. All models share the same test windows and latent actions. We evaluate autoregressive rollouts at horizons $h\in\{1,4,8,15\}$ and report the mean and standard deviation over four training seeds.

\noindent{\textbf{FDM Configurations.}
As detailed in Tab.~\ref{tab:appx:fdms}, all configurations use the same state encoder and dynamics backbone, with only the action-conditioning mechanism varied. For a fair comparison, each FDM contains approximately $18$M parameters.

\paragraph{Detailed Results.}
As summarized in Tab.~\ref{tab:appx:utils}, the detailed utilization scores demonstrate that AT-FDM shows the strongest increase with rollout horizon and achieves the highest average at longer horizons. These results suggest that continuous state-aware action modulation enables the FDM to make more effective utilization of latent actions throughout the rollout.

\begin{table}[t]
\centering
\caption{\textbf{Action-conditioning configurations.} Only the action-conditioning mechanism differs.}
\label{tab:appx:fdms}
\vspace{0.3em}
\setlength{\tabcolsep}{2em}
\renewcommand{\arraystretch}{1.15}
\resizebox{0.88\linewidth}{!}{
\begin{tabular}{ll}
\toprule
\textbf{FDM} & \textbf{Action-conditioning mechanism} \\
\midrule
Add-FDM
& Broadcast action embeddings are added to all patch tokens \\
XAttn-FDM
& Patch queries attend to action embeddings as keys and values \\
Concat-FDM
& Patch and action embeddings are concatenated and jointly projected \\
AdaLN-FDM
& Action-conditioned adaptive modulation of normalized patch tokens \\
AT-FDM
& Continuous state-aware action modulation via evolving action tokens \\
\bottomrule
\end{tabular}}
\end{table}
\vspace{-1em}

\begin{table*}[t]
\centering
\caption{\textbf{The detailed action utilization scores across four datasets.} Values are reported as mean $\pm$ standard deviation over 4 random seeds.}
\label{tab:appx:utils}
\vspace{0.3em}
\setlength{\tabcolsep}{1em}
\renewcommand{\arraystretch}{1.12}
\resizebox{0.8\textwidth}{!}{%
\begin{tabular}{llcccc}
\toprule
\multirow{2}{*}{\textbf{Dataset}}
& \multirow{2}{*}{\textbf{FDM}}
& \multicolumn{4}{c}{\textbf{Utilization (\%) $\uparrow$}} \\
\cmidrule(lr){3-6}
& & @1 & @4 & @8 & @15 \\
\midrule
\multirow{5}{*}{SSv2}
& Add-FDM
& \tabnum{11.7\pm0.2} & \tabnum{13.4\pm0.4} & \tabnum{13.1\pm0.3} & \tabnum{10.8\pm1.3} \\
& Concat-FDM
& \tabnum{11.6\pm0.5} & \tabnum{12.8\pm0.6} & \tabnum{11.9\pm0.4} & \tabnum{2.7\pm8.2} \\
& XAttn-FDM
& \tabnum{11.4\pm0.4} & \tabnum{12.6\pm0.9} & \tabnum{11.7\pm1.5} & \tabnum{9.2\pm2.7} \\
& AdaLN-FDM
& \tabnum{13.1\pm0.5} & \tabnum{14.9\pm0.6} & \tabnum{14.7\pm0.8} & \tabnum{14.2\pm1.3} \\
& AT-FDM
& \bm{\tabnum{15.1\pm0.7}} & \bm{\tabnum{22.8\pm0.6}} & \bm{\tabnum{30.6\pm0.4}} & \bm{\tabnum{41.7\pm0.8}} \\
\midrule
\multirow{5}{*}{RT-1}
& Add-FDM
& \tabnum{22.7\pm0.4} & \tabnum{26.8\pm0.4} & \tabnum{25.2\pm0.4} & \tabnum{20.7\pm0.5} \\
& Concat-FDM
& \tabnum{22.9\pm0.3} & \tabnum{27.3\pm0.2} & \tabnum{25.8\pm0.4} & \tabnum{20.9\pm1.1} \\
& XAttn-FDM
& \tabnum{22.6\pm0.3} & \tabnum{27.0\pm0.5} & \tabnum{25.6\pm1.0} & \tabnum{21.2\pm2.0} \\
& AdaLN-FDM
& \bm{\tabnum{24.6\pm0.2}} & \bm{\tabnum{28.9\pm0.4}} & \tabnum{27.1\pm0.4} & \tabnum{22.1\pm0.4} \\
& AT-FDM
& \tabnum{24.1\pm0.7} & \tabnum{28.6\pm0.5} & \bm{\tabnum{28.4\pm0.5}} & \bm{\tabnum{26.9\pm0.8}} \\
\midrule
\multirow{5}{*}{RECON}
& Add-FDM
& \tabnum{29.6\pm0.5} & \tabnum{35.6\pm1.4} & \tabnum{42.6\pm2.8} & \tabnum{54.7\pm4.3} \\
& Concat-FDM
& \tabnum{29.6\pm0.8} & \tabnum{34.8\pm2.3} & \tabnum{41.0\pm3.5} & \tabnum{51.8\pm3.9} \\
& XAttn-FDM
& \tabnum{29.7\pm0.3} & \tabnum{35.5\pm0.9} & \tabnum{42.7\pm1.8} & \tabnum{54.3\pm2.5} \\
& AdaLN-FDM
& \bm{\tabnum{32.2\pm0.4}} & \tabnum{38.5\pm1.1} & \tabnum{46.0\pm1.9} & \tabnum{58.1\pm5.0} \\
& AT-FDM
& \tabnum{31.1\pm1.5} & \bm{\tabnum{44.2\pm3.2}} & \bm{\tabnum{56.8\pm4.9}} & \bm{\tabnum{70.8\pm5.8}} \\
\midrule
\multirow{5}{*}{LoopNav}
& Add-FDM
& \tabnum{17.8\pm1.5} & \tabnum{20.9\pm2.1} & \tabnum{22.0\pm2.4} & \tabnum{23.9\pm3.1} \\
& Concat-FDM
& \tabnum{18.7\pm1.7} & \tabnum{22.0\pm2.3} & \tabnum{23.2\pm2.5} & \tabnum{25.4\pm2.3} \\
& XAttn-FDM
& \tabnum{17.3\pm0.6} & \tabnum{20.3\pm0.9} & \tabnum{21.5\pm1.0} & \tabnum{23.8\pm1.5} \\
& AdaLN-FDM
& \tabnum{19.9\pm1.7} & \tabnum{23.1\pm2.1} & \tabnum{23.9\pm2.0} & \tabnum{25.4\pm1.6} \\
& AT-FDM
& \bm{\tabnum{20.9\pm1.7}} & \bm{\tabnum{28.1\pm2.7}} & \bm{\tabnum{34.8\pm3.6}} & \bm{\tabnum{45.9\pm3.5}} \\
\midrule
\multirow{5}{*}{Average}
& Add-FDM
& \tabnum{20.4\pm0.6} & \tabnum{24.2\pm0.9} & \tabnum{25.7\pm1.3} & \tabnum{27.5\pm1.9} \\
& Concat-FDM
& \tabnum{20.7\pm0.8} & \tabnum{24.2\pm1.3} & \tabnum{25.5\pm1.5} & \tabnum{25.2\pm2.8} \\
& XAttn-FDM
& \tabnum{20.2\pm0.3} & \tabnum{23.9\pm0.6} & \tabnum{25.4\pm0.9} & \tabnum{27.1\pm1.3} \\
& AdaLN-FDM
& \tabnum{22.5\pm0.3} & \tabnum{26.4\pm0.1} & \tabnum{27.9\pm0.1} & \tabnum{30.0\pm1.0} \\
& AT-FDM
& \bm{\tabnum{22.8\pm1.1}} & \bm{\tabnum{30.9\pm1.7}} & \bm{\tabnum{37.7\pm2.3}} & \bm{\tabnum{46.3\pm2.4}} \\
\bottomrule
\end{tabular}}
\end{table*}

\subsection{Visual Planning}
\label{sec:appx:vp}
\noindent{\textbf{Action Adaptation.}}
We evaluate the model's ability in robotic control on the VP$^2$ benchmark \citep{iclr23_VP2}, following the protocol used in prior works \citep{icml25_AdaWorld,icml26_DiLA}. Since LAMs operate in the latent action space, we first train an action-to-latent MLP for each environment. Specifically, we randomly sample $100$ training trajectories, yielding $3,400$ transition pairs. For each transition, the pretrained IDM extracts the corresponding latent action. The action adapter is a two-layer MLP with SiLU activations. The raw action dimension is $d_a=4$ for RoboSuite and $d_a=5$ for RoboDesk. The pretrained LAM remains frozen during adaptation. We optimize the adapter with SGD for $3,000$ epochs using a learning rate of $0.01$ and a batch size of $10$.

\noindent{\textbf{Joint Finetuning.}}
After obtaining the action-to-latent adapter, we replace the pretrained IDM with the learned MLP and jointly finetune the MLP with the FDM. In this stage, we sample 16-frame windows from the training trajectories and optimize the model for $3,000$ steps using AdamW with a learning rate of $1\times10^{-4}$, a weight decay of $1\times10^{-5}$, and a batch size of 8. The joint finetuning is optimized by $\mathcal{L}_{rec}$ and $\mathcal{L}_{grid}$.

\noindent{\textbf{Visual Planning with VP$^2$.}}
Once finetuning is complete, the LAM is used as the predictive dynamics model within the official VP$^2$ implementation. Each evaluation consists of $100$ trajectories for RoboSuite and $30$ trajectories for RoboDesk, with a maximum trajectory length of $15$ control steps. The planner takes $2$ context frames and harnesses a planning horizon of $10$. At each planning step, MPPI \citep{icra16_MPPI} performs a single optimization using $200$ candidate action sequences for RoboSuite and most RoboDesk tasks. For \emph{Open Slide}, following DiLA \citep{icml26_DiLA}, we increase the number to $800$.


\begin{figure*}[t]
  \centering
   \includegraphics[width=0.45\linewidth]{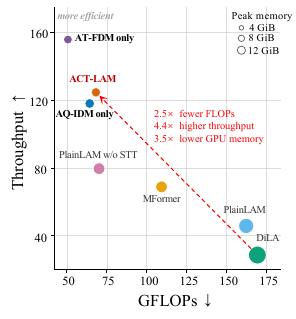}
    \vspace{-0.5em}
    \caption{\textbf{Training efficiency at batch size 1.} 
    The x-axis reports GFLOPs per RGB frame, the y-axis reports throughput in frames/s, and bubble area denotes peak GPU memory. The dashed arrow highlights the efficiency improvement of ACT-LAM over DiLA.}
   \label{fig:appx:efficiency}
   \vspace{-1em}
\end{figure*}

\section{Efficiency Evaluation}
\label{sec:appx:efficiency}
We evaluate the training-time efficiency of the compared LAM architectures under the teacher-forcing paradigm. All measurements use 16-frame $256\times256$ RGB clips, FP32 precision, and an RTX A6000 GPU. FLOPs include the forward and backward passes but exclude the optimizer update. We report GFLOPs per RGB frame, throughput in frames/s, and peak allocated GPU memory.

As shown in Fig.~\ref{fig:appx:efficiency}, the upper-left region corresponds to lower computation and higher throughput. Compared with DiLA, ACT-LAM requires $2.49\times$ fewer GFLOPs, achieves $4.36\times$ higher throughput, and uses $3.47\times$ less peak GPU memory. Under the same hardware and protocol, PlainLAM and DiLA reach out-of-memory at batch size 8, whereas the remaining models run successfully. The detailed results are summarized in Tab.~\ref{tab:appx:efficiency}.

\begin{table*}[t]
\centering
\caption{\textbf{Training efficiency at batch size 1.}}
\label{tab:appx:efficiency}
\vspace{0.3em}
\setlength{\tabcolsep}{4pt}
\renewcommand{\arraystretch}{1.4}
\resizebox{0.85\linewidth}{!}{
\begin{tabular}{lcccc}
\toprule
\textbf{Model}
& \color{gray!80}\textbf{Params. (M)}
& \textbf{GFLOPs} $\downarrow$
& \textbf{Peak memory (GiB)} $\downarrow$
& \textbf{Throughput (frames/s)} $\uparrow$ \\
\midrule
PlainLAM
& \color{gray!80}118.4
& 162.4
& 9.4 
& 45.8 \\
PlainLAM w/o STT
& \color{gray!80}67.3
& 70.1
& 5.7 
& 79.6 \\
MFormer
& \color{gray!80}39.9
& 109.3
& 5.6 
& 68.9 \\
DiLA
& \color{gray!80}123.0
& 169.4
& 13.0 
& 28.6 \\
\midrule
AQ-IDM only
& \color{gray!80}44.7
& 64.3
& 4.0 
& 118.1 \\
AT-FDM only
& \color{gray!80}78.7
& 50.6
& 3.3 
& 155.9 \\
ACT-LAM
& \color{gray!80}54.6
& 68.2
& 3.7 
& 124.7 \\
\bottomrule
\end{tabular}}
\vspace{-0.8em}
\end{table*}

\section{More Analyses}
\label{sec:appx:analyses}

\afterpage{
\begin{figure*}[!t]
  \centering
  \includegraphics[width=1.0\linewidth]{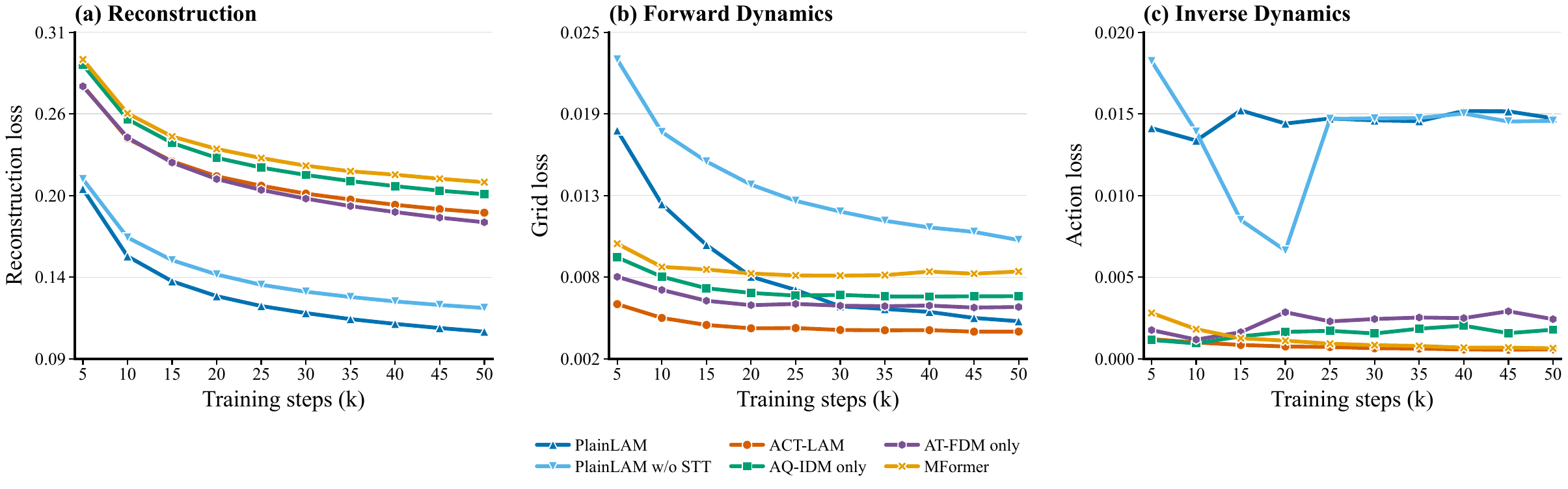}
  \vspace{-1em}
  \caption{\textbf{Detailed validation losses during pretraining.} We report the reconstruction loss, gird loss, and action loss of the compared LAMs throughout $50$k steps. These curves provide a detailed view of the pretraining trends of different LAM architectures.}
  \label{fig:appx:recon_action}
\end{figure*}

\begin{figure}[h]
  \centering
  \includegraphics[width=1.0\linewidth]{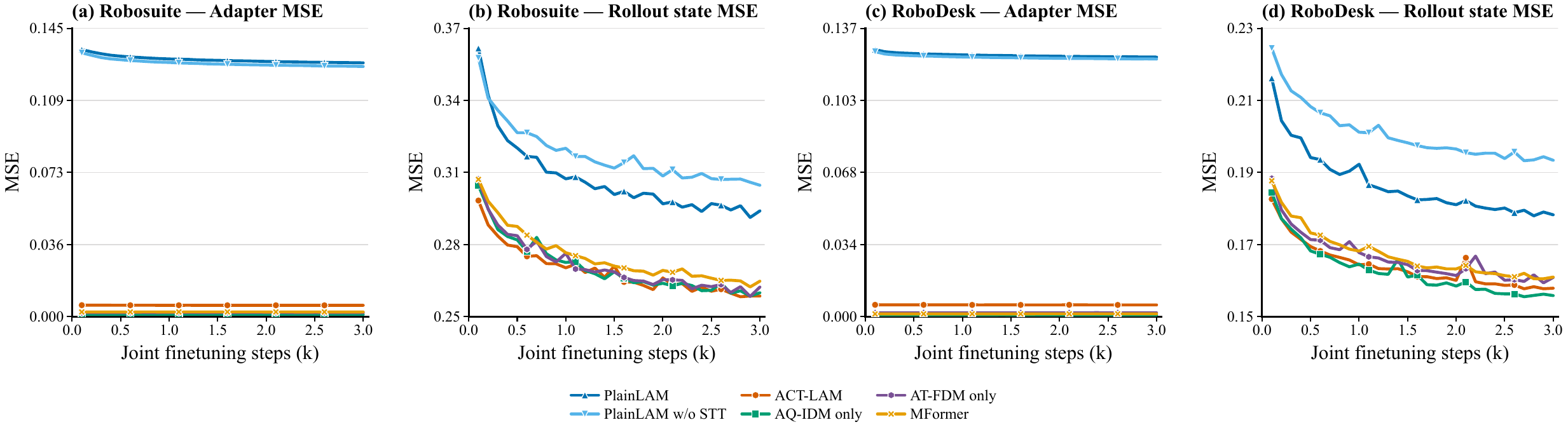}
  \vspace{-1em}
  \caption{\textbf{Detailed validation curves during VP$^2$ joint finetuning.} We report the adapter MSE and rollout state MSE on the Robosuite and RoboDesk benchmarks throughout finetuning. The curves show the adaptation and predictive dynamics trends of the compared LAMs.}
  \label{fig:appx:vp2_finetuning}
\end{figure}

\begin{figure}[h]
  \centering
  \includegraphics[width=0.8\linewidth]{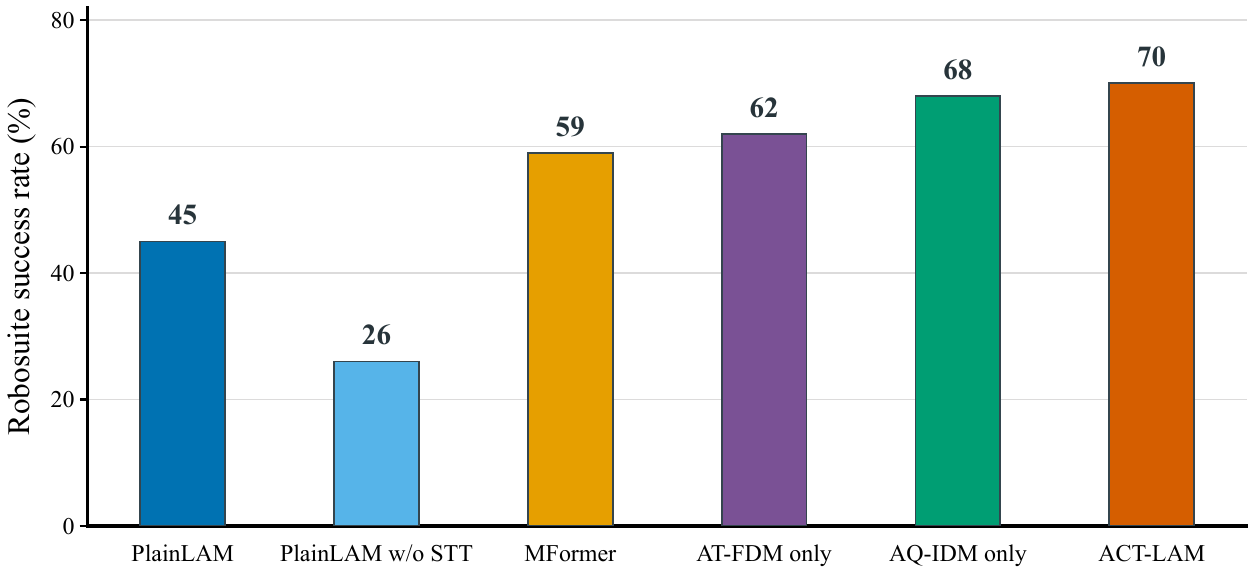}
  \vspace{-1em}
    \caption{\textbf{VP$^2$ ablation.} Success rates of different LAMs on the RoboSuite benchmark. All methods are evaluated under the same protocol. Results are reported using a single random seed due to the high computational cost.}
  \label{fig:appx:vp2_ablation}
\end{figure}

}

\subsection{LAM Pretraining}
\label{sec:appx:lam_pretraining}
As shown in Fig.~\ref{fig:appx:recon_action}, we visualize the validation losses of the compared LAMs over $50$k pretraining steps, including the reconstruction loss, grid loss, and action loss. These losses correspond to the visual reconstruction, latent state reconstruction, and IDM action prediction objectives, respectively. Among these curves, PlainLAM achieves a relatively low reconstruction loss, while ACT-LAM consistently attains lower grid and action losses, indicating more effective latent action learning and improved latent dynamics modeling.

\subsection{VP$^2$ Finetuning}
\label{sec:appx:vp2_finetuning}
We further examine the validation dynamics during VP$^2$ joint finetuning. As shown in Fig.~\ref{fig:appx:vp2_finetuning}, it reports the adapter MSE and rollout state MSE on Robosuite and RoboDesk over 3k finetuning steps. Although \emph{PlainLAM} and \emph{PlainLAM without STT} achieve relatively strong visual reconstruction during pretraining (see Fig.~\ref{fig:appx:recon_action} (a)), they exhibit substantially higher adapter and rollout state errors in joint finetuning. This contrast indicates that the reconstruction quality alone does not guarantee an effective action-conditioned dynamics. In comparison, ACT-LAM generally obtains lower rollout errors, suggesting better compatibility with robot action adaptation. 

\subsection{VP$^2$ Ablation}
\label{sec:appx:vp2_ablation}
As illustrated in Fig.~\ref{fig:appx:vp2_ablation}, we evaluate the individual contributions of AQ-IDM and AT-FDM on the VP$^2$ benchmark. The full ACT-LAM achieves the highest success rate, while both single-component models remain competitive, demonstrating that AQ-IDM and AT-FDM provide complementary gains. Given the high computational cost of VP$^2$ evaluation, we report these ablation results using a single random seed.

\subsection{VP$^2$ Video Prediction}
\label{sec:appx:vp2_video}
As shown in Fig.~\ref{fig:appx:vp2_visual}, we visualize representative VP$^2$ rollout sequences on RoboSuite and RoboDesk, comparing Simulator (top) and ACT-LAM (bottom). For each trajectory, we uniformly sample observations with a stride of 2. Notably, ACT-LAM reaches the desired state in fewer steps in several cases, whereas Simulator requires additional control steps. These examples suggest that ACT-LAM can produce temporally efficient action sequences, providing qualitative evidence complementary to the aggregate success rate results.

\afterpage{
\begin{figure*}[t]
  \centering
  \includegraphics[width=1.0\linewidth]{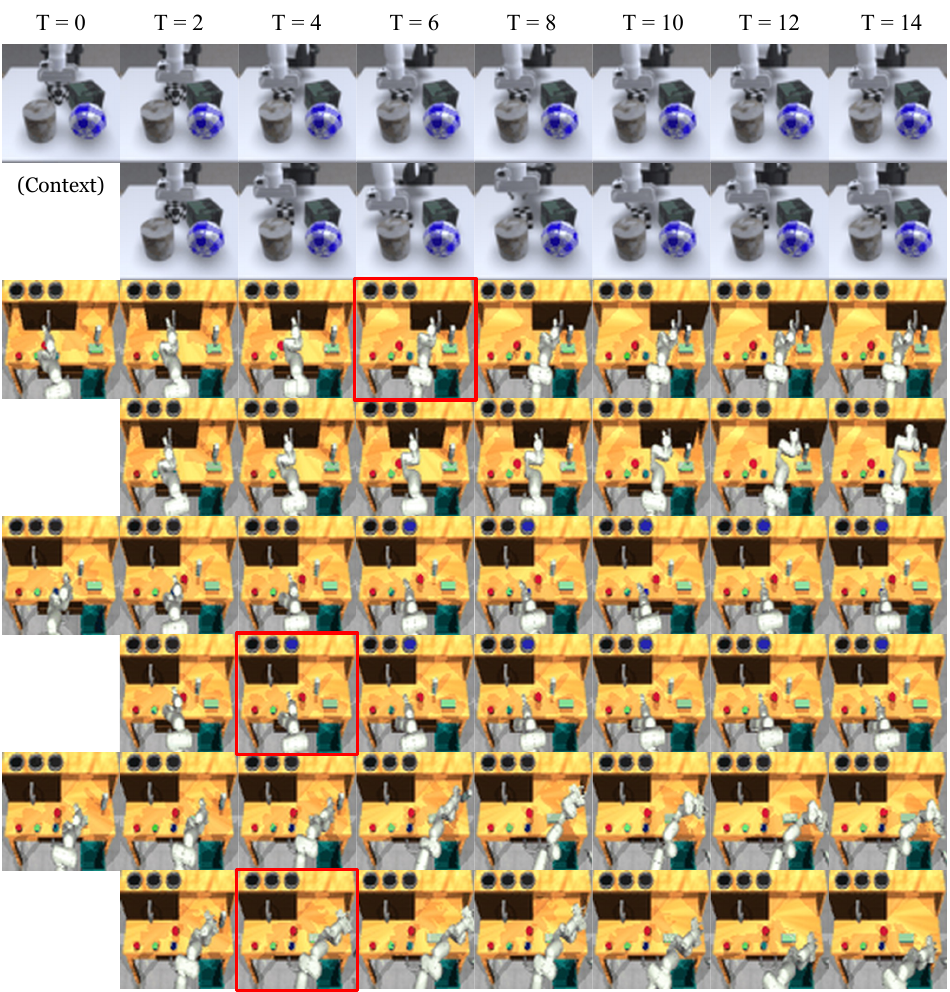}
  \vspace{-1em}
    \caption{\textbf{Visualization of video prediction results.} Predicted rollouts generated by Simulator (top) and ACT-LAM (bottom) on VP$^2$ tasks including robosuite and RoboDesk.}
  \label{fig:appx:vp2_visual}
\end{figure*}
}
\end{document}